\PassOptionsToPackage{table}{xcolor}
\documentclass[]{selfevolagent}

\usepackage{microtype}
\microtypesetup{expansion=false}
\usepackage{graphicx}
\usepackage{subcaption}
\usepackage{booktabs} 
\usepackage{hyperref}

\usepackage{amsmath}
\usepackage{amssymb}
\usepackage{mathtools}
\usepackage{amsthm}

\usepackage{latexsym}
\usepackage[T1]{fontenc}
\usepackage[utf8]{inputenc}
\usepackage{microtype}
\usepackage{amsmath}
\usepackage{booktabs}
\usepackage{multirow}
\usepackage{amssymb}
\usepackage{balance}
\usepackage{subcaption}

\usepackage{pifont}
\usepackage{makecell}
\usepackage{filecontents}
\usepackage{bbm}
\usepackage{pgfplots}
\usetikzlibrary{patterns}
\usepackage[inline,shortlabels]{enumitem}
\usepackage{natbib}
\usepackage{csquotes}
\usepackage{hyperref}
\usepackage{url}
\usepackage{svg}
\usepackage{bm}
\usepackage{graphics}
\usepackage{graphicx}
\usepackage{array}
\usepackage[english]{babel}
\usepackage{textcomp}
\usepackage{latexsym}
\usepackage{siunitx}  
\usepackage[normalem]{ulem}
\useunder{\uline}{\ul}{}
\usepackage{array}
\usepackage{titletoc}
\usepackage{tikz}
\usepackage{arydshln}
\usepackage[most]{tcolorbox}
\usepackage{algorithm}
\usepackage{algorithmic}
\usepackage{fontawesome5}
\usepackage{wrapfig}
\usetikzlibrary{tikzmark}
\makeatletter
\newcommand*\myfontsize{%
\@setfontsize\myfontsize{7}{8}%
}
\makeatother
\newcommand{\mytextbox}[2]{\tikzmarknode[draw=#1,thick,inner sep=2pt]{test}{\myfontsize #2}}
\definecolor{geminiBlue}{HTML}{8E8ED7}
\definecolor{qwenBlue}{HTML}{78A2E0}

\definecolor{myred}{rgb}{0.7, 0.3, 0.0}
\definecolor{myblue}{HTML}{0a41b8}
\definecolor{mygreen}{HTML}{056b34}
\definecolor{mypurple}{HTML}{5d1e8b}

\newcommand{\blue}[1]{\mytextbox{myblue}{\textbf{\textcolor{myblue}{#1}}}}

\newcommand{\purple}[1]{\mytextbox{mypurple}{\textbf{\textcolor{mypurple}{#1}}}}

\newcommand{\githubicon}{\raisebox{-1.5pt}{\includegraphics[height=1.03em]{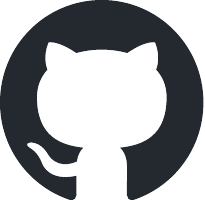}}}
\newcommand{\huggingfaceicon}{\raisebox{-1.5pt}{\includegraphics[height=0.96em]{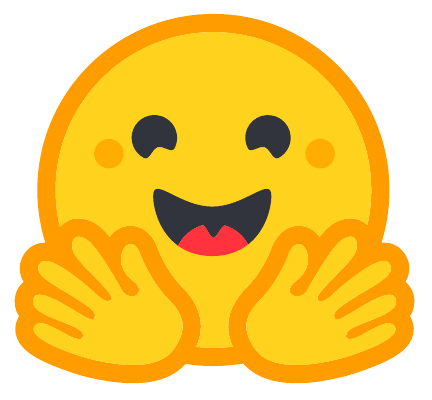}}}

\newcommand{\modelicon}[2]{\raisebox{-0.2\height}{\includegraphics[height=#1]{fig/icons/#2}}}

\theoremstyle{plain}

\theoremstyle{definition}

\theoremstyle{remark}

\usepackage[textsize=tiny]{todonotes}
\usepackage{xcolor}

\definecolor{startBlue}{HTML}{1628a7}  
\definecolor{endPurple}{HTML}{8b16aa}
\newcommand{\elegantOmni}{%
\textcolor{startBlue}{\textbf{O}}
\textcolor{startBlue!60!endPurple}{\textbf{m}}
\textcolor{endPurple!60!startBlue}{\textbf{n}}
\textcolor{endPurple}{\textbf{i}}
}

\title{\elegantOmni GAIA: Towards Native Omni-Modal AI Agents}
\author[1,2*]{Xiaoxi Li}
\author[2]{Wenxiang Jiao}
\author[2]{Jiarui Jin}
\author[3]{Haoxuan Li}
\author[2,3*]{Hao Wang}
\author[2,4*]{Shijian Wang}
\author[1]{Guanting Dong}
\author[1]{\protect\\Jiajie Jin}
\author[2,5*]{Yinuo Wang}
\author[2]{Yuan Lu}
\author[1]{Ji-Rong Wen}
\author[1\dagger]{Zhicheng Dou}
\author[3\dagger]{Zhouchen Lin}

\affiliation[1]{Renmin University of China}
\affiliation[2]{Xiaohongshu Inc.}
\affiliation[3]{Peking University}
\affiliation[4]{Southeast University}
\affiliation[5]{Tsinghua University}
\contribution[*]{Work done during internship at Xiaohongshu}
\contribution[\dagger]{Corresponding author}

\metadata[Contact]{xiaoxi\_li@ruc.edu.cn, dou@ruc.edu.cn}
\metadata[\githubicon~Code \& Demo]{\href{https://github.com/RUC-NLPIR/OmniGAIA}{https://github.com/RUC-NLPIR/OmniGAIA}}
\metadata[\huggingfaceicon~Dataset \& Model]{\href{https://huggingface.co/collections/RUC-NLPIR/omnigaia}{https://huggingface.co/collections/RUC-NLPIR/omnigaia}}
\metadata[\raisebox{-0.2ex}{\scalebox{0.9}{\faChartLine}}~Leaderboard]{\href{https://huggingface.co/spaces/RUC-NLPIR/OmniGAIA-LeaderBoard}{https://huggingface.co/spaces/RUC-NLPIR/OmniGAIA-LeaderBoard}}

\abstract{Human intelligence naturally intertwines omni-modal perception---spanning vision, audio, and language---with complex reasoning and tool usage to interact with the world. However, current multi-modal LLMs are primarily confined to bi-modal interactions (e.g., vision-language), lacking the unified cognitive capabilities required for general AI assistants. To bridge this gap, we introduce \textbf{OmniGAIA}, a comprehensive benchmark designed to evaluate omni-modal agents on tasks necessitating deep reasoning and multi-turn tool execution across video, audio, and image modalities. Constructed via a novel omni-modal event graph approach, OmniGAIA synthesizes complex, multi-hop queries derived from real-world data that require cross-modal reasoning and external tool integration. Furthermore, we propose \textbf{OmniAtlas}, a native omni-modal foundation agent under tool-integrated reasoning paradigm with active omni-modal perception. Trained on trajectories synthesized via a hindsight-guided tree exploration strategy and \textit{OmniDPO} for fine-grained error correction, OmniAtlas effectively enhances the tool-use capabilities of existing open-source models. This work marks a step towards next-generation native omni-modal AI assistants for real-world scenarios.}

\begin{document}
\maketitle

\section{Introduction}

Human intelligence seamlessly intertwines language, vision, and audio with long-horizon reasoning and tool use to understand the world and take actions. Building general-purpose AI assistants therefore requires models that can \emph{jointly} perceive across modalities, reason over long contexts, and interact with external tools for verification and knowledge acquisition. Yet, despite rapid progress, multimodal LLM research is still dominated by bi-modal settings (e.g., vision--language or audio--language), which limits their ability to handle truly interwoven real-world modalities.

Emerging omni-modal foundation models (e.g., Qwen3-Omni~\citep{Qwen3-Omni}) have begun to unify richer modalities, but most efforts primarily emphasize perception, leaving \emph{tool-integrated, agentic reasoning} underexplored. Evaluation also lags behind: existing benchmarks are largely bi-modal and perception-centric (e.g., OmniBench~\citep{OmniBench}, WorldSense~\citep{WorldSense}, UNO-Bench~\citep{UNO-Bench}), and thus do not adequately measure multi-hop omni-modal reasoning and multi-turn external tool use with verifiable open-form answers.

\begin{figure*}[!t]
\centering
\includegraphics[width=0.96\linewidth]{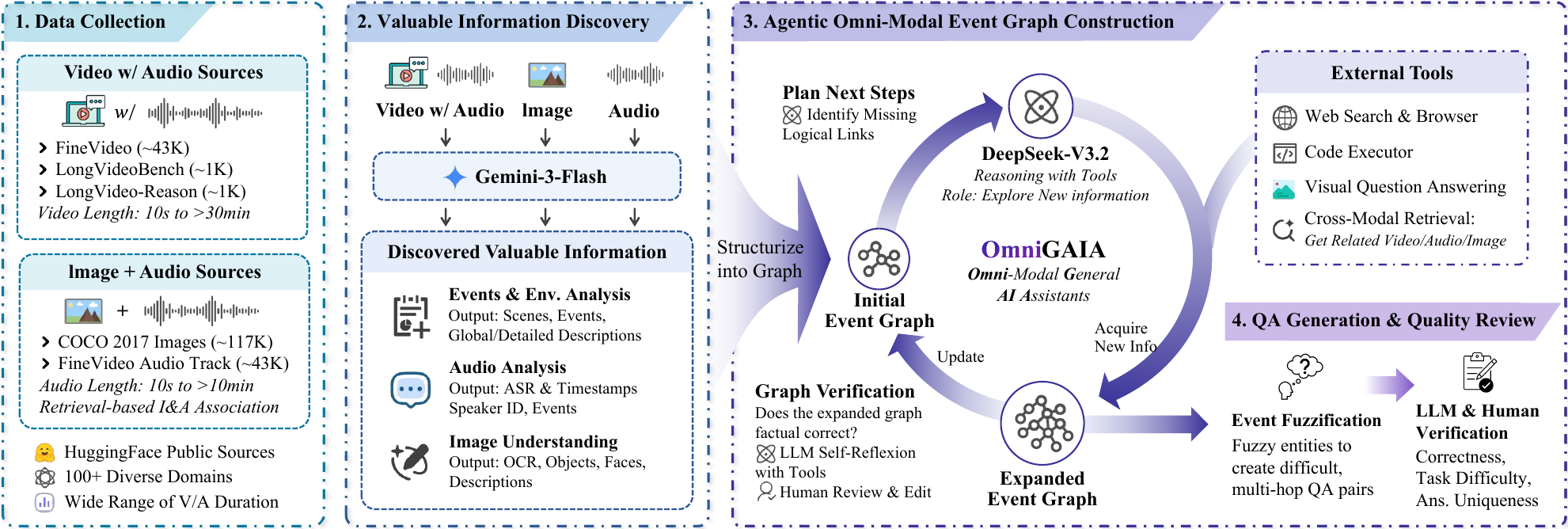}
\caption{\textbf{OmniGAIA construction pipeline.} From video w/ audio and image + audio data, we mine key signals, build and expand a tool-augmented event graph, and generate LLM and human-verified multi-hop QA via event fuzzification.}
\label{fig:omnigaia_construction}
\end{figure*}

To bridge this gap, we introduce \textbf{OmniGAIA, a challenging benchmark for \emph{native omni-modal agents}}. OmniGAIA comprises 360 tasks across 9 real-world domains, covering both video-with-audio and image+audio settings, and explicitly requires multi-turn tool use (e.g., web search/browsing and code) to produce verifiable \emph{open-form} answers. To structure time-aligned multimodal cues and tool-related evidence for multi-hop reasoning, OmniGAIA is constructed via an \emph{omni-modal event-graph-driven} pipeline (\Cref{fig:omnigaia_construction}): (1) we collect data and mine fine-grained signals from raw media; (2) we build an initial event graph that connects cross-modal entities/events and relations; (3) we expand the graph with \emph{next-hop} evidence via cross-modal retrieval and external tools; and (4) we fuzzify key nodes/edges to generate multi-hop QA, followed by LLM screening and human verification for solvability and uniqueness.

Beyond benchmarking, we propose \textbf{OmniAtlas, a native omni-modal foundation agent} following the \emph{Tool-Integrated Reasoning} (TIR) paradigm that naturally interleaves reasoning and tool calls. OmniAtlas further supports \emph{active omni-modal perception} to selectively ``look'' or ``listen'' to the segments/regions in long media without blanket downsampling. For training, we synthesize high-quality tool-integrated trajectories via \emph{hindsight-guided tree exploration}, perform trajectory-level supervised learning, and further propose \textit{\textbf{OmniDPO}} for fine-grained error correction.

\begin{figure*}[!t]
\centering
\includegraphics[width=0.99\linewidth]{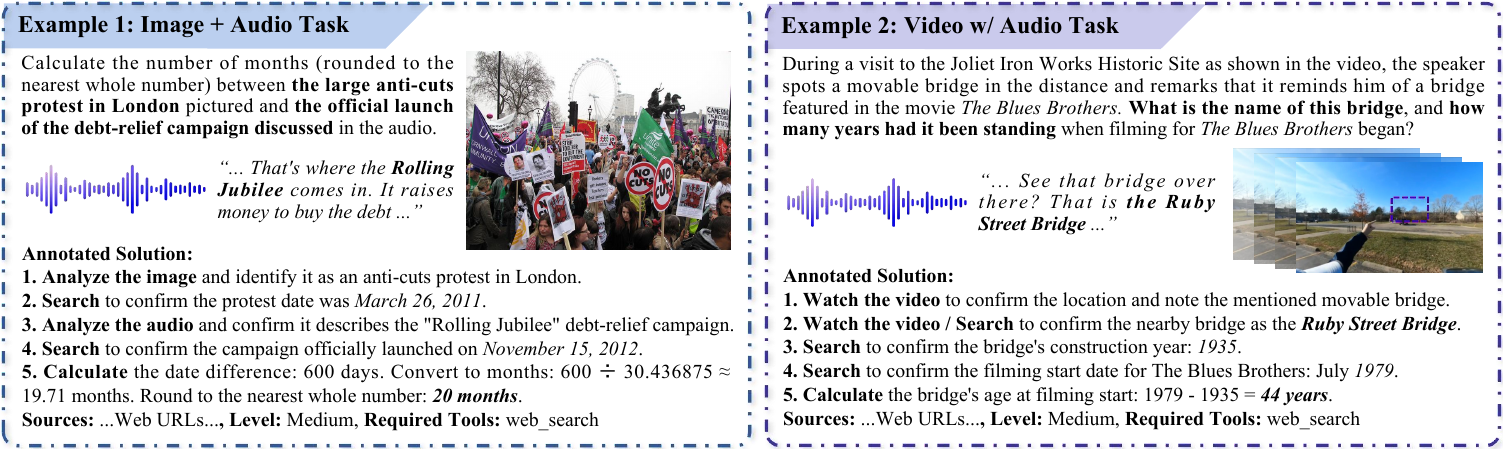}
\caption{\textbf{Short examples from OmniGAIA.} Two illustrative (image + audio and video w/ audio) instances showing omni-modal evidence integration and multi-step tool use (e.g., web search) to derive a verifiable final answer.}
\label{fig:examples}
\end{figure*}

\begin{table*}[t]
\centering
\fontsize{7.65pt}{9.2pt}\selectfont
\caption{\textbf{Comparison of OmniGAIA with existing benchmarks.} ``Video'', ``Image'', and ``Audio'' denote the supported modalities. ``MC'' indicates multiple-choice questions, whereas ``Open'' indicates open-form generation.}
\label{tab:comparison}
\setlength{\tabcolsep}{3pt}
\begin{tabular}{l cccccc cccc}
\toprule
\makecell[l]{{\bfseries Benchmark}} & {\bfseries Video} & {\bfseries Image} & {\bfseries Audio} & \makecell{{\bfseries Multi-hop} \\ {\bfseries Reasoning}} & \makecell{{\bfseries External} \\ {\bfseries Tools}} & \makecell{{\bfseries Multi-} \\ {\bfseries Domain}} & \makecell{{\bfseries Video} \\ {\bfseries Duration}} & \makecell{{\bfseries Audio} \\ {\bfseries Duration}} & \makecell{{\bfseries Answer} \\ {\bfseries Type}} & \makecell{{\bfseries Qwen3-Omni} \\ {\bfseries Accuracy}} \\
\midrule
GAIA~\citep{GAIA} & \ding{55} & \ding{51} & \ding{55} & \ding{51} & \ding{51} & \ding{51} & - & - & Open & - \\
AV-Odyssey~\citep{AV-Odyssey} & \ding{55} & \ding{51} & \ding{51} & \ding{55} & \ding{55} & \ding{51} & - & 3-364 s & MC & - \\
OmniBench~\citep{OmniBench} & \ding{55} & \ding{51} & \ding{51} & \ding{55} & \ding{55} & \ding{51} & - & 0.6-31 s & MC & 58.4 \\
Daily-Omni~\citep{Daily-Omni} & \ding{51} & \ding{55} & \ding{51} & \ding{55} & \ding{55} & \ding{55} & 30/60 s & 30/60 s & MC & 75.8 \\
WorldSense~\citep{WorldSense} & \ding{51} & \ding{55} & \ding{51} & \ding{55} & \ding{55} & \ding{51} & 15-656 s & 15-656 s & MC & 54.0 \\
OmniVideoBench~\citep{OmniVideoBench} & \ding{51} & \ding{55} & \ding{51} & \ding{51} & \ding{55} & \ding{51} & 4-1955 s & 4-1955 s & MC & 38.4 \\
VideoDR~\citep{VideoDR} & \ding{51} & \ding{55} & \ding{55} & \ding{51} & \ding{51} & \ding{51} & 10-288 s & - & Open & 37.0 \\
UNO-Bench~\citep{UNO-Bench} & \ding{51} & \ding{51} & \ding{51} & \ding{51} & \ding{55} & \ding{51} & 0.7-641 s & 1-641 s & MC/Open & 42.1/37.1 \\
\midrule
{\bfseries OmniGAIA (Ours)} & \ding{51} & \ding{51} & \ding{51} & \ding{51} & \ding{51} & \ding{51} & 20-2352 s & 20-657 s & Open & 13.3 \\
\bottomrule
\end{tabular}
\end{table*}

Experiments show that OmniGAIA is highly challenging: the strongest proprietary model (Gemini-3-Pro) reaches 62.5 Pass@1, while an open-source baseline (Qwen3-Omni) achieves 13.3. Our OmniAtlas recipe substantially improves open models (e.g., Qwen3-Omni: 13.3$\rightarrow$20.8). Further analyses of fine-grained error types, tool-use behaviors, and perception strategies expose key limitations of current methods and point to promising directions for future omni-modal agents.
Our main contributions are:
\begin{itemize}[leftmargin=*,nosep]
\item We introduce \textbf{OmniGAIA, a challenging benchmark for native omni-modal agents}, featuring video/image/audio inputs, multi-domain coverage, multi-hop reasoning, multi-turn tool use, and open-form answers.
\item We propose a scalable \textbf{Event-graph-driven Construction Pipeline} that systematically synthesizes hard yet solvable tasks from real-world data.
\item We present \textbf{OmniAtlas, a native omni-modal foundation agent} with active perception and tool-integrated reasoning, together with a practical training recipe (trajectory synthesis, supervised learning, and \textbf{OmniDPO}) that significantly improves open-source backbones.
\item We provide comprehensive evaluations and analyses, including category-wise results, fine-grained error breakdowns, and tool-use behavior studies that highlight key bottlenecks for omni-modal agents.
\end{itemize}

\section{Related Work}

\subsection{Omni-Modal Foundation Models and Benchmarks}
Building on advances in pure-text~\citep{llama3}, vision-language~\citep{GPT-4o}, and audio-language~\citep{Qwen2-Audio} foundation models, recent omni-modal models seek to unify text, vision, and audio within a single LLM backbone. A common approach adopts a unified tokenization-and-projection interface that maps heterogeneous visual and acoustic inputs into a shared token space~\citep{Qwen3-Omni,NEXUS-O,NExT-OMNI,OmniVinci}. Concurrent work further strengthens omni-modal reasoning behaviors~\citep{Omni-R1,M3-Agent}.
For evaluation, existing benchmarks (e.g., OmniBench~\citep{OmniBench}, WorldSense~\citep{WorldSense}, Daily-Omni~\citep{Daily-Omni}, UNO-Bench~\citep{UNO-Bench}) largely emphasize short audios/videos and perception-centric tasks, leaving long-horizon reasoning and tool-integrated agency underexplored. 

\subsection{Autonomous Agents}
LLM-driven autonomous agents tackle real-world tasks by reasoning and acting through external tools that interface with their environment~\citep{wang_llm_agent_survey,LLM-Agent-Survey}. 
Existing approaches broadly fall into workflow-based paradigms~\citep{yao2022react,Plan-and-Solve,CodeAct,AgentFlow} and native agentic reasoning methods~\citep{ToRL,ToolRL,ReTool,VerlTool}, and have shown strong performance on text-only tasks. Moving beyond text, recent studies investigate vision-language agents for multimodal web search~\citep{MM-BrowseComp,MMSearch-R1,WebWatcher}, long-form video understanding~\citep{VideoAgent,Deep-Video-Discovery,VideoARM}, and GUI navigation~\citep{OSWorld,AppAgent,Mobile-Agent}. However, \emph{omni-modal} foundation agents that natively fuse audio, vision, and language while performing long-horizon agentic reasoning remain underexplored. Such capabilities are essential for building general-purpose AI assistants in real-world scenarios.

\section{OmniGAIA: Benchmarking Omni-Modal General AI Assistants}
OmniGAIA is a benchmark of challenging \emph{omni-modal agentic} tasks designed to stress-test unified perception over vision, audio, and language, together with long-horizon reasoning and multi-turn tool use in realistic scenarios. 

\begin{figure*}[!t]
\centering
\includegraphics[width=0.99\linewidth]{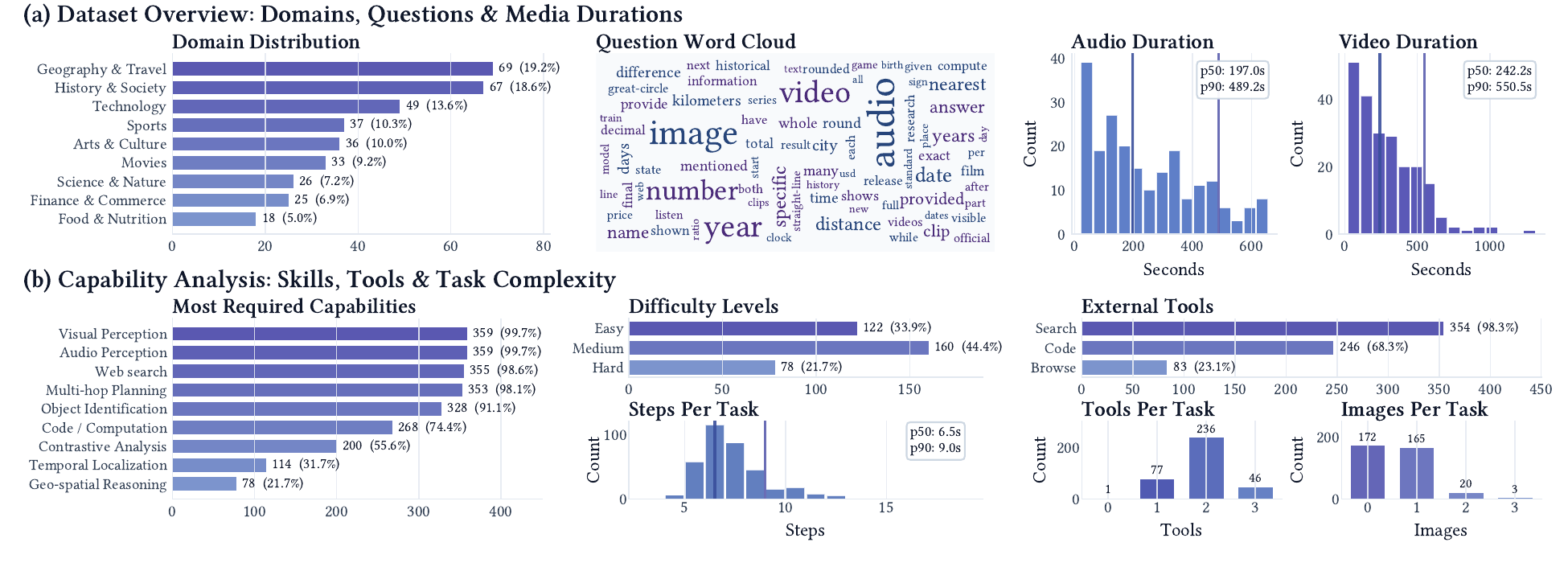}
\vspace{-0.2cm}
\caption{\textbf{OmniGAIA statistics.} This figure presents a detailed breakdown of domain distributions, required capabilities, and task attributes, underscoring the complex demands placed on omni-modal perception, reasoning, and tool utilization.}
\label{fig:statistics}
\end{figure*}

\subsection{Data Collection}
To reflect the complexity of real-world omni-modal interactions, we construct OmniGAIA from two complementary settings: (i) \textbf{video with audio}, and (ii) \textbf{image + audio} pairs.

For the video setting, we aggregate high-quality videos from multiple sources to ensure diversity in both content and duration. We include \textbf{FineVideo}~\citep{FineVideo} (43K videos spanning broad domains; average length 4 minutes). To evaluate long-context reasoning, we further incorporate \textbf{LongVideoBench}~\citep{LongVideoBench} (\(\sim\)1K videos) and \textbf{LongVideo-Reason}~\citep{LongVILA} (\(\sim\)1K videos), both containing videos around 10 minutes.

For the image + audio setting, we use audio tracks from FineVideo to provide diverse acoustic environments, and draw images from \textbf{COCO 2017}~\citep{COCO}, which contains 122K complex everyday-scene images with object detection and segmentation annotations.

\subsection{Discovering Valuable Information}
We employ a strong omni-modal model (Gemini-3-Flash) to extract fine-grained, time-aware signals from each modality for task construction. For videos, we split each video into clips of at most 60 seconds to capture subtle temporal details, and generate both clip-level and full-video descriptions covering scenes, events, and non-speech ambient sounds. For audio, we run timestamped automatic speech recognition (ASR), speaker diarization, and audio event detection; we also tag non-speech acoustic environments (e.g., street, indoor, stadium, nature) and produce global audio summaries. For images, we apply optical character recognition (OCR), recognize objects and faces, and generate a holistic caption to summarize visual content.

\subsection{Omni-modal Event Graph Construction}
To reliably synthesize complex multi-hop tasks, we build an \emph{omni-modal event graph} that structures the discovered information into an explicit graph for each sample. This graph serves as the backbone of our event-graph-driven construction pipeline, enabling systematic evidence expansion and controllable information fuzzification for QA generation.

Using the extracted information, we leverage a strong reasoning agent DeepSeek-V3.2 to automatically build an event graph that represents entities/events and their cross-modal relations. Importantly, real-world logic is rarely a simple linear chain; it often exhibits branching (one-to-many), cascading (sequential), and mixed topologies. The graph representation captures such structures and supports reliable synthesis of logically consistent, challenging tasks.

\subsection{Agentic Omni-modal Event Graph Expansion}
Given an initial event graph, we introduce \textbf{Agentic Event Graph Expansion} to proactively discover missing evidence and create tasks that truly require cross-modal association and external tool use. Following the Tool-Integrated Reasoning (TIR) paradigm, we use a strong reasoning model (DeepSeek-V3.2) as an exploration agent that searches for \emph{next-hop valuable information} and links it back to the graph.

\textbf{Functionality for the event exploration agent.} We equip the agent with a set of omni-modal and external tools:
\begin{itemize}[leftmargin=*,nosep]
\item \textbf{Cross-modal sources linking:} The agent can call \texttt{search\_related\_\{video/audio/image\}\_info} to retrieve context-related multi-modal materials from our database. This is crucial when current graph is insufficient for a tightly-coupled multi-hop question. For the image + audio setting, we pre-retrieve the initial related audios candidates to encourage explicit cross-modal reasoning.
\item \textbf{Web knowledge integration:} With \texttt{web\_search} and \texttt{page\_browser}, the agent can retrieve top web pages and read detailed content, enabling time-sensitive, verifiable external knowledge beyond the original media.
\item \textbf{External visual exploration:} Using \texttt{web\_image\_search} and \texttt{visual\_question\_answering}, the agent can search web images and query their content, expanding task construction to scenarios requiring external visual evidence.
\item \textbf{Computation:} The \texttt{code\_executor} tool supports complex computations (e.g., arithmetic, statistics), enabling tasks that require reliable multi-step numerical reasoning.
\end{itemize}
During task generation, these tools are embedded in the prompting interface, and the agent autonomously decides whether and how to invoke them to expand the information boundary of the current graph, producing complex QA pairs enriched with next-hop evidence.

\subsection{QA Pairs Generation via Event Fuzzification}
To convert expanded graphs into truly challenging tasks, we propose \textbf{QA generation via event fuzzification}. Directly querying a graph node often reduces to trivial fact lookup. Instead, we select specific nodes/edges along long reasoning paths and apply \emph{fuzzy entities} (e.g., replacing a specific entity with its type, or masking key attributes) to mask or abstract key information. This forces models to traverse the full logical path and integrate multi-source, multi-modal evidence to derive a unique answer.

\subsection{Quality Inspection}
To ensure rigor and high quality, we apply an inspection pipeline with \emph{LLM screening} and \emph{human verification}, with an optional difficulty expansion step in between.

\textbf{1. LLM screening:} We form a review committee with DeepSeek-V3.2 and Gemini-3-Pro to automatically evaluate each QA pair across multiple criteria: (i) naturalness and clarity of the question; (ii) indispensability of omni-modal perception and tool use (filtering out unimodal or trivial cases); and (iii) answer correctness and uniqueness.

\textbf{2. Difficulty expansion:} For preliminarily qualified samples, we optionally increase difficulty by linking additional data sources, mining deeper evidence, or introducing more complex computation steps.

\textbf{3. Human review:} Finally, we invite three graduate-level computer science reviewers to verify each QA pair against the underlying media. They check question soundness, annotation correctness, and answer correctness/uniqueness, and fix minor issues to ensure each test case is reliably solvable and high-quality.

\subsection{Statistics}
As shown in \Cref{fig:statistics}, OmniGAIA comprises 360 omni-modal agentic tasks across 9 real-world domains, intentionally designed to stress \emph{long-horizon} perception and \emph{tool-integrated} reasoning. Tasks often require grounding evidence from both vision and audio over minutes-long media, planning multi-step solution paths, and verifying or extending information via external tools (primarily web search, and occasionally code/computation). The statistics highlight that performance hinges not only on native perception, but also on reliable multi-hop planning and effective tool use under long contexts.

\begin{figure*}[!t]
\centering
\includegraphics[width=0.95\linewidth]{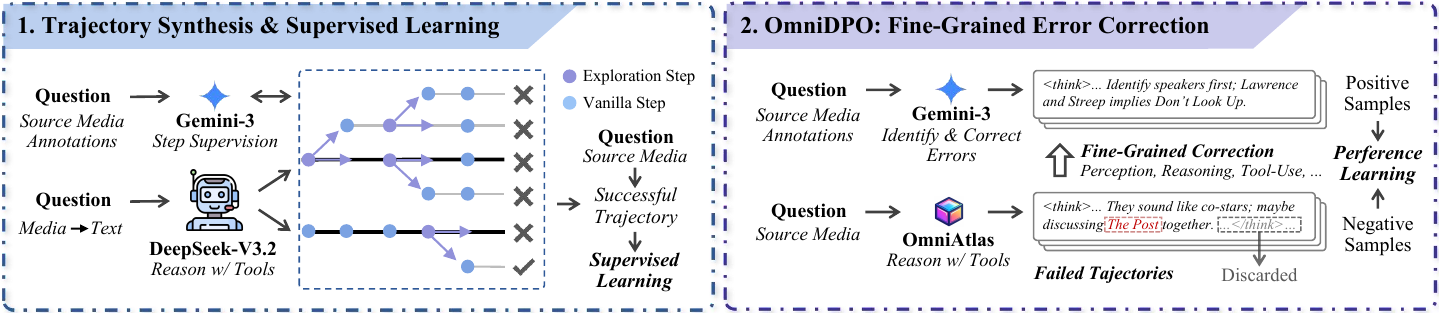}
\caption{\textbf{OmniAtlas training strategy.} Left, we synthesize tool-integrated trajectories via step-level supervision and guided tree exploration, selecting successful runs for supervised fine-tuning; right, \textbf{OmniDPO} locates the first error in a failed trajectory and generates a corrected prefix, forming positive/negative preference pairs for fine-grained correction.}
\label{fig:omniatlas_train}
\end{figure*}

\section{OmniAtlas: Omni-Modal Foundation Agent}
\label{sec:OmniAtlas}
In this section, we introduce \textbf{OmniAtlas}, a native omni-modal foundation agent that unifies vision, audio, and language perception with long-horizon reasoning and autonomous tool use. To overcome the key weaknesses of current open-source omni-modal models in perception and tool-integrated reasoning, we present a comprehensive training and optimization recipe.

\subsection{Autonomous Tool-Integrated Reasoning}
To enable OmniAtlas to acquire external knowledge and handle complex tasks, we integrate tools like \emph{web search}, \emph{page browser}, and \emph{code executor}. The agent adopts a \emph{tool-integrated reasoning} paradigm, autonomously switching between internal reasoning and tool usage as needed.

Formally, we define an agent trajectory as $\tau = [(s_t, a_t, o_t)]_{t=0}^{T}$, where $s_t$ denotes the reasoning thought at step $t$, $a_t$ the action (either a tool call or a final response), and $o_t$ the observation returned by the tool (empty if no tool is invoked). The model generates the next thought and action conditioned on the interaction history:
\begin{equation}
  p_\theta(\tau \mid \mathbf{x}) = \prod\nolimits_{t=0}^T p_\theta(s_t, a_t \mid \mathbf{x}, s_{<t}, a_{<t}, o_{<t})
\end{equation}
Here, $\mathbf{x}$ denotes the user instruction and omni-modal inputs. When tool-call tokens are detected, generation is paused, the corresponding tool is executed, and the returned observation $o_t$ is appended to the context so the model can continue. This design preserves intermediate reasoning states and supports coherent long-horizon problem solving, aligning with the tool-integrated generation philosophy of DeepSeek-V3.2~\citep{DeepSeek-V3.2}.

\paragraph{Active Omni-Modal Perception.} 
For long videos or high-resolution images, naively ingesting all media is token-expensive and often requires aggressive downsampling that can discard critical details~\citep{LLaVA-OneVision}. To mitigate this, OmniAtlas supports \emph{active} omni-modal perception: the agent can selectively request the specific segments or regions it needs via operations such as \texttt{read\_video(video\_id, t\_start, t\_end)}, \texttt{read\_audio(audio\_id, t\_start, t\_end)}, and \texttt{read\_image(image\_ids, crop\_box)}. When invoked, the corresponding raw media content is loaded into the model context, enabling ``look-where-needed'' perception without blanket downsampling.

\subsection{Trajectory Synthesizing via Guided Tree Exploration}
Our preliminary experiments on OmniGAIA show that open-source omni-modal models still lag behind in both omni-modal perception and tool-integrated reasoning. To internalize these capabilities, we synthesize high-quality agent trajectories via a two-stage pipeline: (i) we use Gemini-3-Flash to convert raw multi-modal inputs into detailed textual descriptions; (ii) we then generate tool-augmented solution trajectories using \textit{\textbf{Hindsight-Guided Tree Exploration}}.

Concretely, since proprietary Gemini models do not expose raw reasoning traces, we use strong reasoning agent DeepSeek-V3.2 to synthesize tool-integrated trajectories. Starting from the root state, we sample $k=3$ candidate continuations (reasoning + tool actions) at each step and use a verifier (Gemini-3-Flash), conditioned on the ground-truth answer, to prune incorrect or redundant branches; we keep only successful trajectories for training (\Cref{fig:omniatlas_train}).

\begin{table*}[!t]
\centering
\caption{\textbf{Main results on the OmniGAIA benchmark.} The Pass@1 metric is reported for all tasks. Best and second-best scores are highlighted in \textbf{bold} and \underline{underlined} respectively, shown separately for proprietary and open-source models.}
\label{tab:omnigaia_results}
\setlength\tabcolsep{4.2pt}
\fontsize{8.2pt}{11.4pt}\selectfont
\begin{tabular}{l c ccccccccc ccc c}
\toprule
\multirow{2}[2]{*}{{\bfseries Method}} &
\multirow{2}[2]{*}{{\bfseries \# Params}} &
\multicolumn{9}{c}{{\bfseries Category-Wise Breakdown}} &
\multicolumn{3}{c}{{\bfseries Difficulty Levels}} &
\multirow{2}[2]{*}{{\bfseries Overall}} \\
\cmidrule(lr){3-11} \cmidrule(lr){12-14}
  & & Geo. & Tech. & Hist. & Fin.
    & Sport & Art & Movie & Sci. & Food
    & Easy & Med. & Hard &  \\
\midrule
\rowcolor[HTML]{f0f0f0}
\multicolumn{15}{c}{{\bfseries\itshape Proprietary Omni-Modal Models}} \\
\modelicon{1.1em}{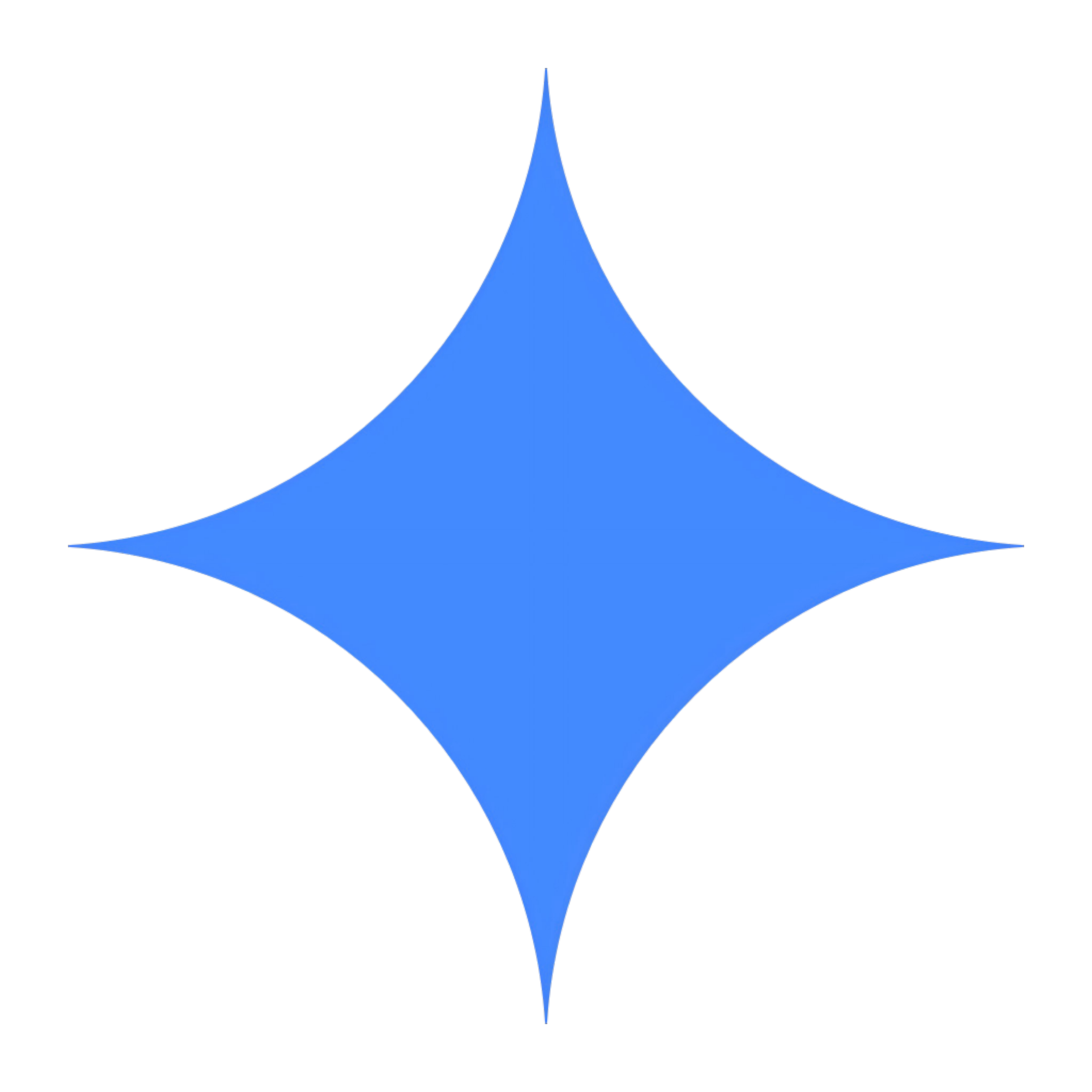}\,Gemini-2.5-Flash-Lite
& - & 5.8 & 8.2 & 14.9 & 4.0 & 10.8 & 8.3 & 6.1 & 3.9 & 11.1 & 9.8 & 8.1 & 7.7 & 8.6 \\
\modelicon{1.1em}{gemini.pdf}\,Gemini-2.5-Pro
& - & 23.2 & 28.6 & 32.8 & 20.0 & 32.4 & 41.7 & 42.4 & 26.9 & 33.3 & 41.8 & 26.9 & 21.8 & 30.8 \\
\modelicon{1.1em}{gemini.pdf}\,Gemini-3-Flash
& - & \underline{50.7} & \underline{57.1} & \underline{44.8} & \underline{48.0}
& \underline{59.5} & {\bfseries 55.6} & {\bfseries 54.6} & \underline{38.5} & \underline{61.1}
& \underline{67.2} & \underline{46.9} & \underline{37.2} & \underline{51.7} \\
\modelicon{1.1em}{gemini.pdf}\,Gemini-3-Pro
& - & {\bfseries 65.2} & {\bfseries 59.2} & {\bfseries 62.1} & {\bfseries 72.0}
& {\bfseries 78.4} & \underline{52.8} & \underline{48.5} & {\bfseries 42.3} & {\bfseries 88.9}
& {\bfseries 78.7} & {\bfseries 61.9} & {\bfseries 38.5} & {\bfseries 62.5} \\
\midrule
\rowcolor[HTML]{f0f0f0}
\multicolumn{15}{c}{{\bfseries\itshape Open-Source Omni-Modal Models}} \\
\modelicon{1.0em}{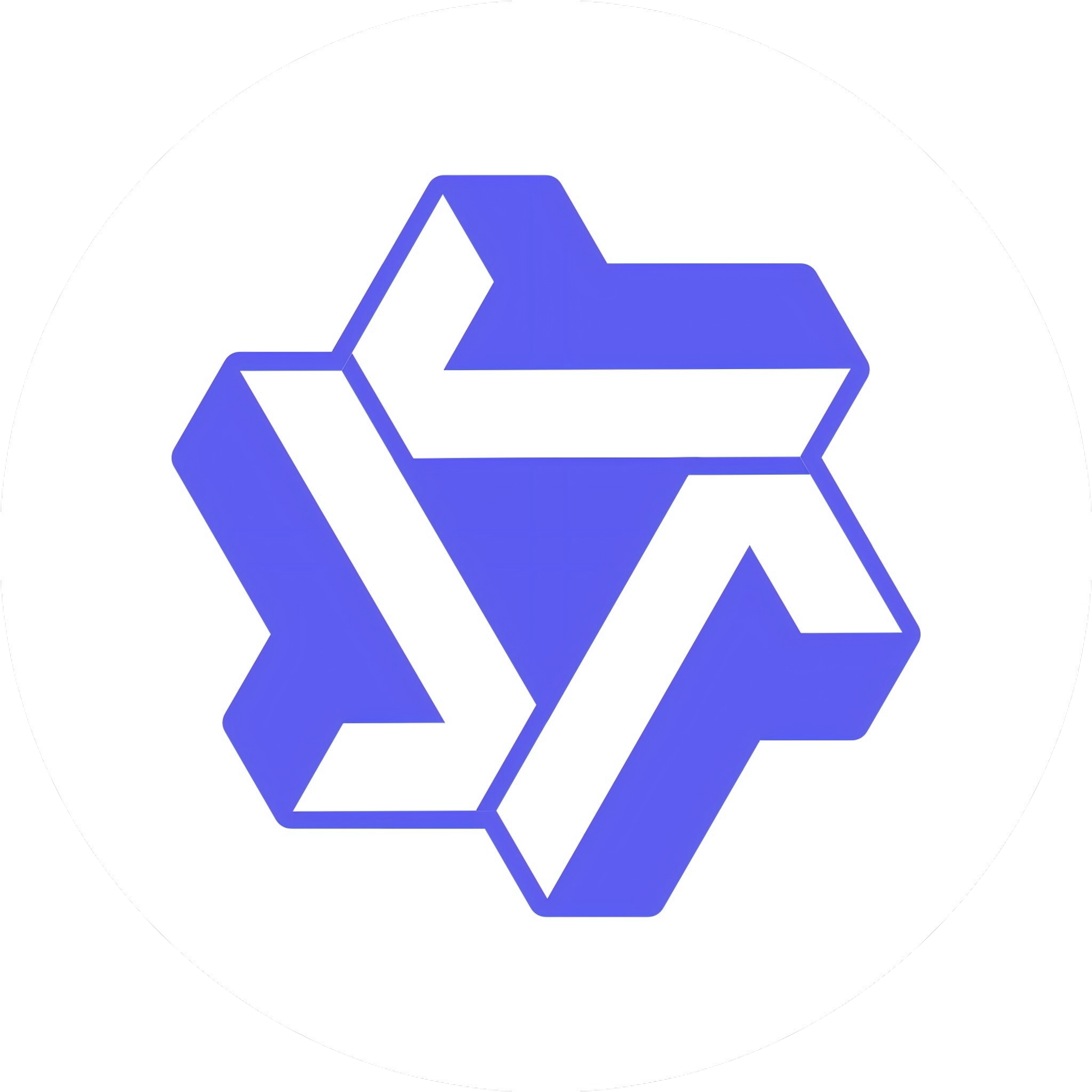}~Qwen-2.5-Omni
& 3B & 0.0 & 2.0 & 4.5 & 0.0 & 0.0 & 0.0 & 0.0 & 3.9 & 0.0 & 1.6 & 1.9 & 0.0 & 1.4 \\
\modelicon{1.0em}{qwen.pdf}~Qwen-2.5-Omni
& 7B & 1.5 & 4.1 & 7.5 & 4.0 & 0.0 & 2.8 & 0.0 & 7.7 & 5.6 & 8.2 & 1.3 & 1.3 & 3.6 \\
\modelicon{0.9em}{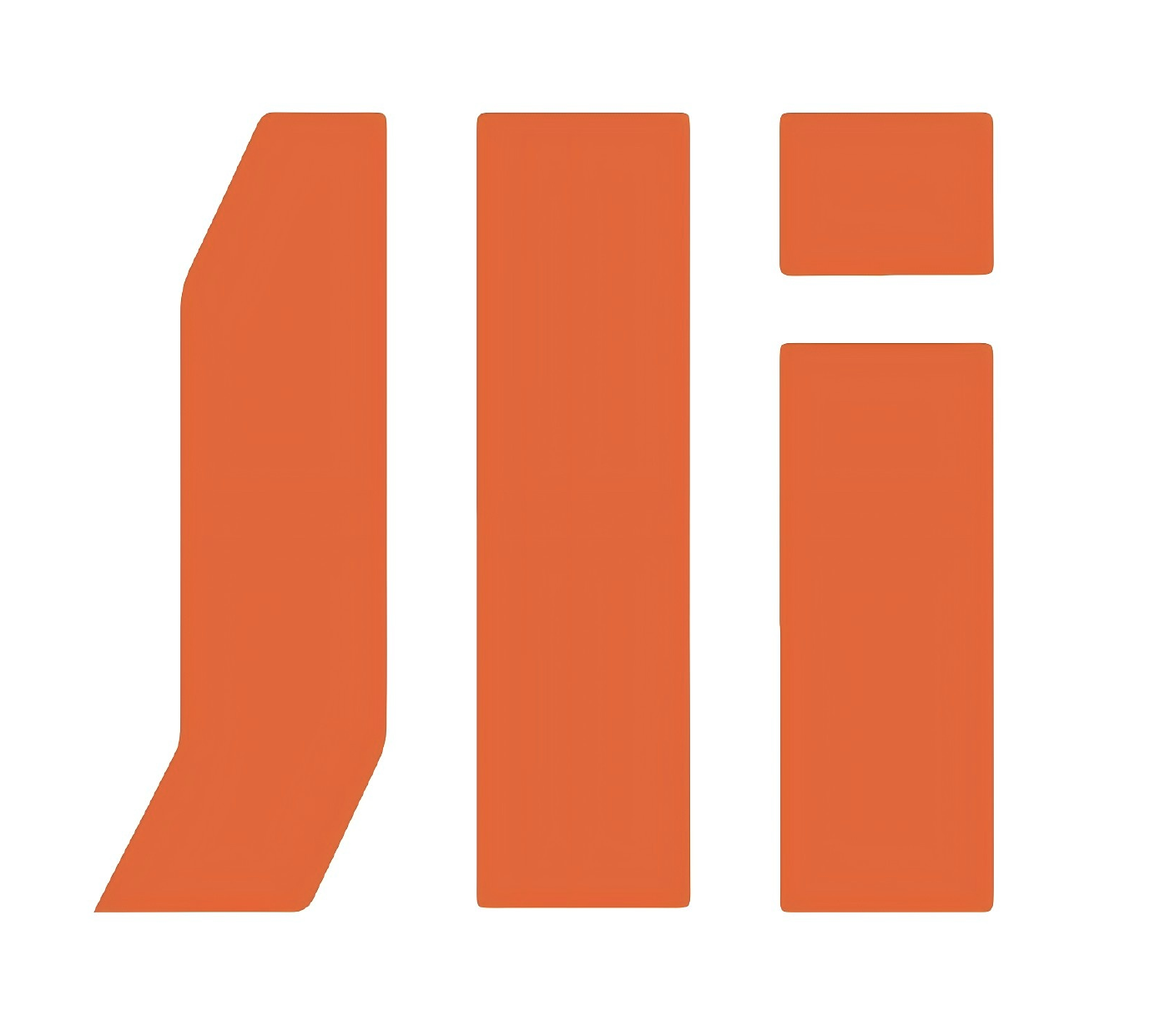}~Baichuan-Omni-1.5
& 8B & 2.9 & 4.1 & 3.0 & 4.0 & 2.7 & 0.0 & 3.0 & 3.8 & 0.0 & 4.9 & 2.5 & 0.0 & 2.8 \\
\modelicon{0.9em}{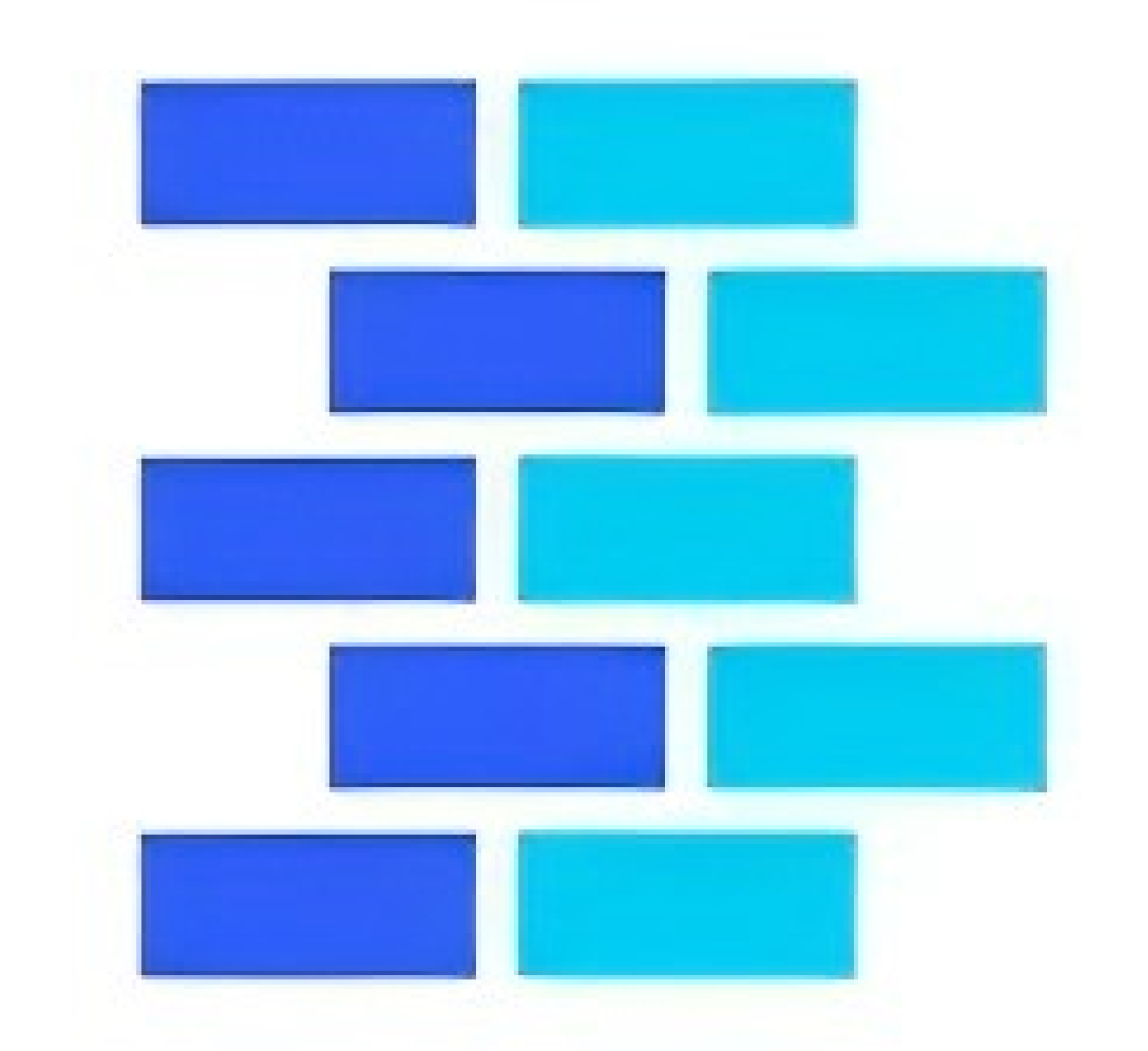}~MiniCPM-O-2.6
& 8B & 2.9 & 2.0 & 1.5 & 0.0 & 2.7 & 8.3 & 3.0 & 3.8 & 5.6 & 3.3 & 2.5 & 3.8 & 3.1 \\
\modelicon{1.0em}{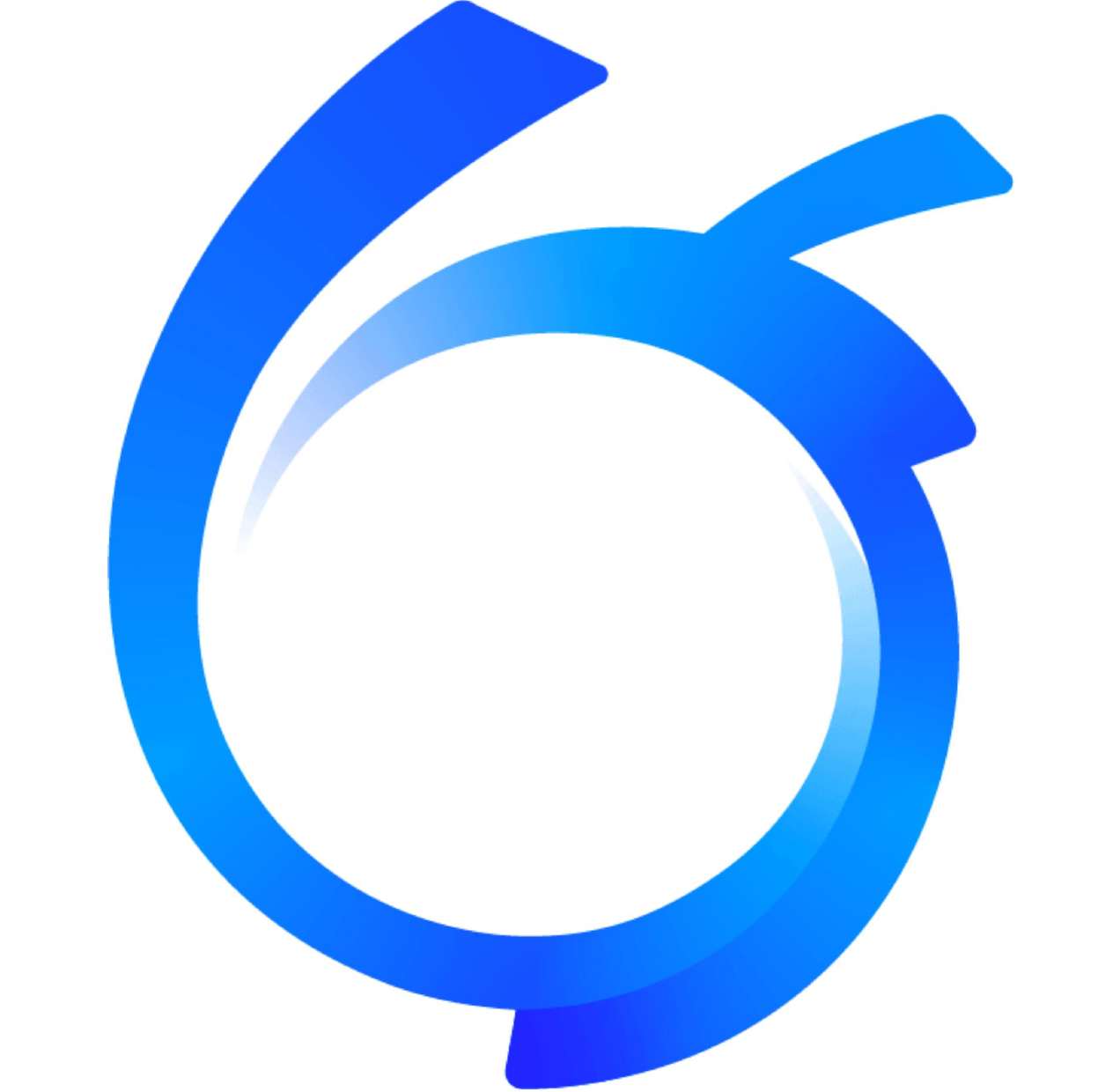}~Ming-Lite-Omni-1.5
& 20B-A3B & 2.9 & 6.1 & 1.5 & 4.0 & 5.4 & 2.8 & 6.1 & 7.7 & 5.6 & 4.9 & 3.8 & 2.6 & 3.9 \\
\modelicon{1.0em}{qwen.pdf}~Qwen-3-Omni
& 30B-A3B & \underline{8.7} & 14.3 & 11.9 & \underline{28.0} & 10.8 & 13.9 & \underline{9.1} & {\bfseries 15.4} & \underline{22.2} & 19.7 & 10.6 & {\bfseries 9.0} & \underline{13.3} \\
\modelicon{1.0em}{ming.pdf}~Ming-Flash-Omni
& 100B-A6B & 5.8 & 8.2 & 10.4 & 12.0 & 8.1 & 5.6 & 6.1 & \underline{11.5} & 11.1 & 12.3 & 7.5 & 3.8 & 8.3 \\
\modelicon{1.0em}{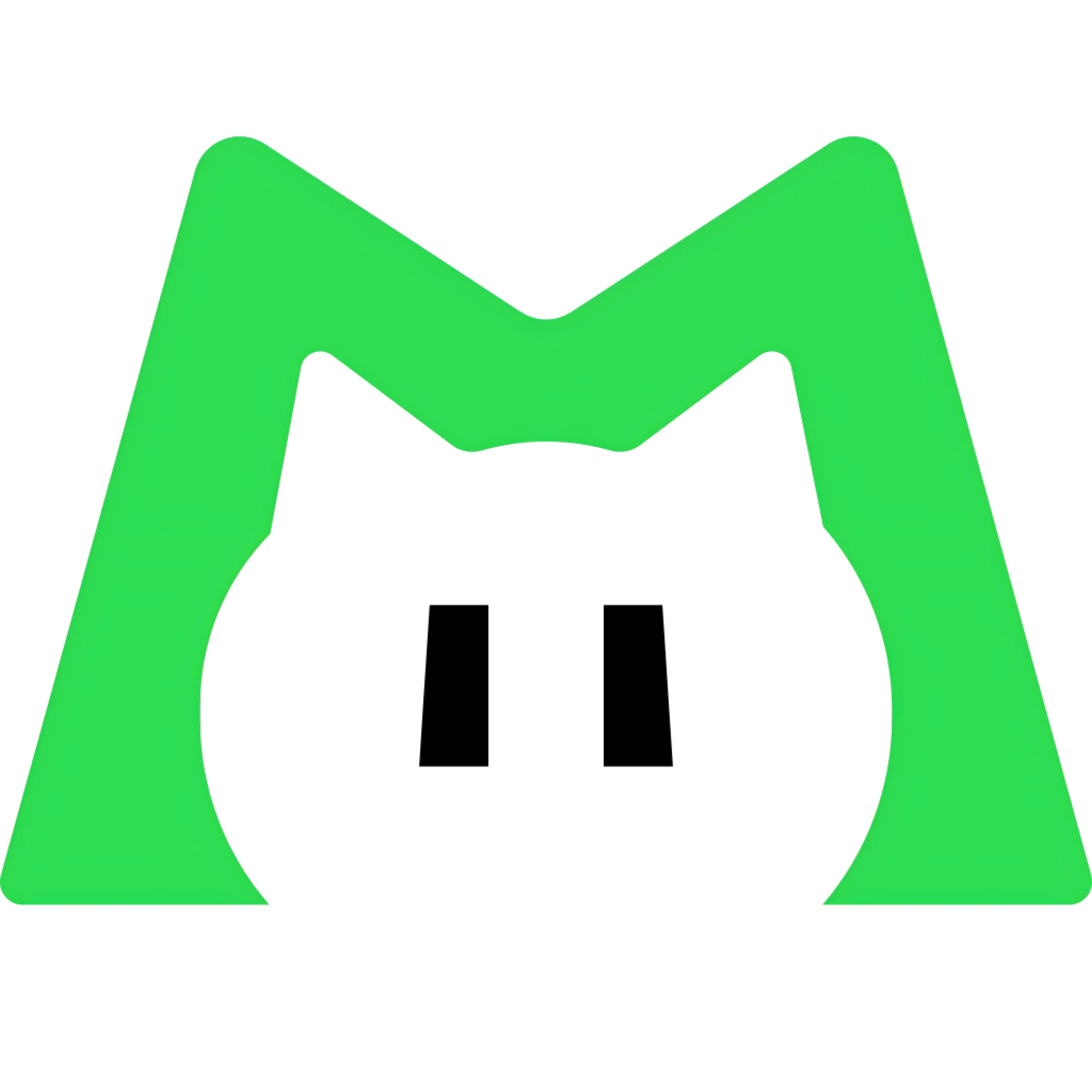}\,\,LongCat-Flash-Omni
& 560B-A27B & \underline{8.7} & 10.2 & 16.4 & 12.0 & 10.8 & 8.3 & 6.1 & \underline{11.5} & 16.7 & 16.4 & 9.4 & \underline{6.4} & 11.1 \\
\hdashline
\rowcolor[HTML]{f0edff}
\modelicon{0.97em}{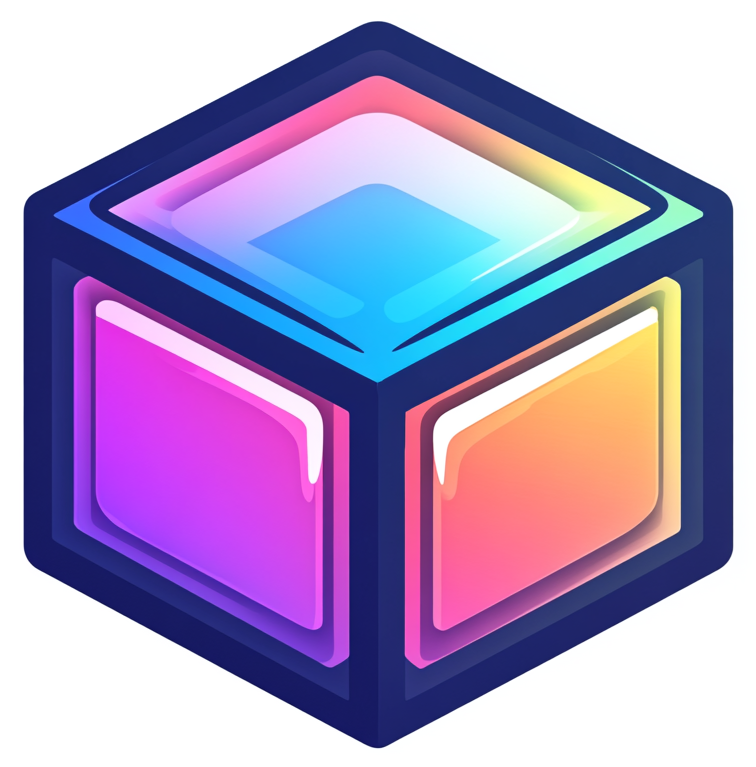}\,\,OmniAtlas-Qwen-2.5
& 3B & 4.4 & 12.2 & \underline{16.7} & 4.0 & \underline{16.2} & 11.1 & 3.0 & \underline{11.5} & 11.1 & 13.9 & 10.0 & 5.1 & 10.3 \\
\rowcolor[HTML]{f0edff}
\modelicon{0.97em}{omniagent_icon.png}\,\,OmniAtlas-Qwen-2.5
& 7B & \underline{8.7} & \underline{18.4} & 16.4 & 4.0 & \underline{16.2} & {\bfseries 22.2} & 3.0 & 7.7 & \underline{22.2} & \underline{22.1} & \underline{11.3} & 3.9 & \underline{13.3} \\
\rowcolor[HTML]{f0edff}
\modelicon{0.97em}{omniagent_icon.png}\,\,OmniAtlas-Qwen-3
& 30B-A3B & {\bfseries 10.1} & {\bfseries 30.6} & {\bfseries 29.9} & {\bfseries 32.0}
& {\bfseries 18.9} & \underline{16.7} & {\bfseries 12.1} & \underline{11.5} & {\bfseries 27.8}
& {\bfseries 31.1} & {\bfseries 18.8} & {\bfseries 9.0} & {\bfseries 20.8} \\
\bottomrule
\end{tabular}
\end{table*}

\subsection{Trajectory-Level Supervised Fine-Tuning}
\label{sec:sft}
We perform \emph{trajectory-level} supervised fine-tuning (SFT) to teach the model effective perception, reasoning, and tool-use behaviors. We use standard teacher forcing, but apply \emph{masked} supervision: we compute loss only on tokens generated by the agent (reasoning and tool-call tokens), while masking out tool observations to prevent memorizing environment feedback.

Let the input sequence be $\mathbf{y} = [y_1, y_2, \dots, y_L]$ with a mask $\mathbf{m} \in \{0, 1\}^L$, where $m_i=1$ iff $y_i$ belongs to the agent's thoughts or actions. The masked SFT objective is:
\begin{equation}
  \mathcal{L}_{\text{SFT}}(\theta) = - \frac{1}{\sum\nolimits_{i=1}^L m_i} \sum\nolimits_{i=1}^{L} m_i \log p_\theta(y_i \mid y_{<i}, \mathbf{x})
\end{equation}
This encourages the model to learn \emph{how to think and act} without fitting the noisy tool observation tokens.

\subsection{OmniDPO: Fine-Grained Error Correction}
Omni-modal agentic tasks require multiple tightly-coupled capabilities (e.g., visual/audio perception, reasoning, and tool use), and full-trajectory SFT alone is often insufficient to correct fine-grained mistakes. We propose \textbf{OmniDPO}, which performs preference optimization on \emph{fine-grained segments} aligned with failure modes, including perception, reasoning, tool use or other specific types of errors.

Specifically, we let the SFT model explore on the training set. For each failed trajectory, Gemini-3-Flash (with access to the annotated solution and answer) identifies the \emph{first} erroneous step and generates a corrected prefix up to that point. This approach enables the training process to concentrate on rectifying a single error per optimization. We denote the original (incorrect) prefix as $\tau_{\text{lose}}$ and the corrected prefix as $\tau_{\text{win}}$, and optimize a masked DPO objective:
\begin{equation}
\mathcal{L}_{\text{DPO}}(\pi_\theta, \pi_{\text{ref}}) = - \mathbb{E}_{(\tau_{\text{win}}, \tau_{\text{lose}}) \sim \mathcal{D}} \Big[ \log \sigma \Big(  \beta \log \frac{\pi_\theta(\tau_{\text{win}})}{\pi_{\text{ref}}(\tau_{\text{win}})} - \beta \log \frac{\pi_\theta(\tau_{\text{lose}})}{\pi_{\text{ref}}(\tau_{\text{lose}})} \Big) \Big]
\end{equation}
Here $\pi_{\text{ref}}$ is a reference policy (typically the SFT model). As in  \Cref{sec:sft}, we compute log-probabilities only on agent-generated tokens, focusing correction on the specific module where the error appears.

\section{Experiments}

\subsection{Experimental Settings}

\paragraph{Evaluation}
We employ LLM-as-a-Judge based on DeepSeek-V3.2~\citep{DeepSeek-V3.2} to evaluate answer equivalence, considering that answers may appear in diverse forms. Pass@1 is reported, where a trial is considered correct if the model's final answer is judged equivalent to the ground truth. The judging prompt is detailed in Appendix~\ref{app:instructions}. All models are provided with the same external tools, including web search, browser, and code executor.

\paragraph{Models}
We evaluate omni-modal foundation models: proprietary models Gemini-2.5-[Flash-Lite, Pro]~\citep{Gemini-2.5} and Gemini-3-[Flash, Pro]~\citep{gemini-3}; and open-source models Qwen2.5-Omni-[3B,7B]~\citep{Qwen2.5-Omni}, Qwen3-Omni-30B-A3B-Thinking~\citep{Qwen3-Omni}, Baichuan-Omni-1.5~\citep{Qwen2.5-Omni}, MiniCPM-O-2.6~\citep{MiniCPM-V}, Ming-Lite-Omni-1.5~\citep{Ming-Omni}, Ming-Flash-Omni~\citep{Ming-Flash-Omni}, and LongCat-Flash-Omni~\citep{LongCat-Flash-Omni}.

\begin{figure*}[!t]
\centering
\includegraphics[width=1\linewidth]{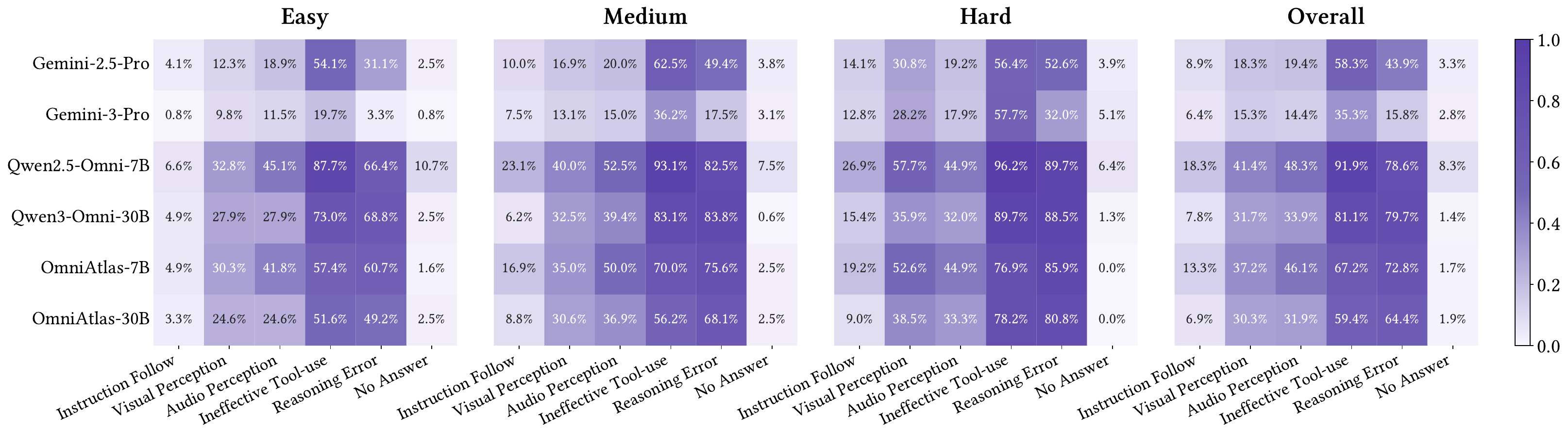}
\caption{\textbf{Fine-grained error analysis.} These heatmaps illustrate the frequency of specific error types—including failures in instruction following, visual/audio perception, tool usage, reasoning, and absence of an answer—across six different models.}
\label{fig:error_analysis_heatmap}
\end{figure*}

\subsection{Main Results}
\Cref{tab:omnigaia_results} summarizes the Pass@1 performance on OmniGAIA under the unified tool setting. The benchmark proves highly challenging: while the state-of-the-art proprietary model, Gemini-3-Pro, achieves 62.5, the strongest open-source baseline, Qwen-3-Omni, reaches only 13.3.

\textit{\textbf{(1) Substantial proprietary--open gap:}} A stark performance disparity exists between Gemini-3-Pro and Qwen-3-Omni ($\sim$4.7$\times$, 62.5 vs. 13.3). This underscores the critical need for advancements in both native omni-modal perception and robust tool-integrated reasoning within the open-source community.

\textit{\textbf{(2) Scaling parameters alone is insufficient:}} Merely increasing model size yields diminishing returns. For instance, the massive LongCat-Flash-Omni (560B) underperforms the smaller Qwen-3-Omni (30B) (11.1 vs. 13.3). This suggests that agentic capabilities—specifically tool-use policies—rather than raw parameter count, are the primary bottleneck.

\textit{\textbf{(3) OmniAtlas delivers consistent improvements:}} Our approach significantly boosts Qwen-3-Omni from 13.3 to 20.8 (+7.5 absolute). Notably, the gains are even more pronounced on smaller backbones (e.g., Qwen-2.5-Omni-7B improves $\sim$3.7$\times$ from 3.6 to 13.3), demonstrating the efficacy of OmniAtlas in unlocking agentic potential across varying model sizes.

\textit{\textbf{(4) Hard tasks remain the main challenge:}} Performance degrades sharply as task difficulty increases (e.g., Gemini-3-Pro drops from 78.7 on Easy to 38.5 on Hard). While OmniAtlas improves performance on Easy and Medium tasks, the ``Hard'' subset—requiring deep multi-hop reasoning—remains a formidable challenge, highlighting significant opportunities for future research.

\subsection{Fine-Grained Error Analysis}
\Cref{fig:error_analysis_heatmap} breaks down fine-grained error types by difficulty. 

\textit{\textbf{(1) Tool-use and reasoning failures predominate:}} Ineffective tool usage and reasoning errors represent the most prevalent failure modes (35.3\%--91.9\% and 15.8\%--79.7\%, respectively), significantly outpacing instruction-following issues (6.4\%--18.3\%) and ``No Answer'' cases (1.4\%--8.3\%).

\textit{\textbf{(2) Hard tasks reveal cascading failure modes:}} On hard tasks, open-source models exhibit near-saturated tool misuse ($\sim$90\%--96\%) alongside high reasoning error rates ($\sim$80\%--90\%). This suggests that initial failures in evidence acquisition via tools propagate downstream, inevitably leading to reasoning collapse.

\textit{\textbf{(3) Proprietary models demonstrate superior robustness:}} Gemini-3-Pro significantly outperforms Qwen-3-Omni, exhibiting much lower error rates in visual/audio perception (15.3\%/14.4\% vs. 31.7\%/33.9\%) and particularly in tool-use/reasoning (35.3\%/15.8\% vs. 81.1\%/79.7\%), reflecting its more mature planning and verification capabilities.

\textit{\textbf{(4) OmniAtlas enhances tool policy, yet perception remains a bottleneck:}} While OmniAtlas effectively reduces tool misuse (e.g., 81.1\%$\rightarrow$59.4\%) and reasoning errors (79.7\%$\rightarrow$64.4\%), visual and audio perception errors remain high ($\sim$30\%--50\%). This indicates that the fundamental perception capability of omni-modal foundation models is a persistent bottleneck requiring further attention. Representative success/failure trajectories are analyzed in Appendix~\ref{app:case_study}.

\newpage

\subsection{Tool Call Distribution Analysis}
\begin{wrapfigure}{r}{0.52\linewidth}
\vspace{-1\baselineskip}
\centering
\includegraphics[width=\linewidth]{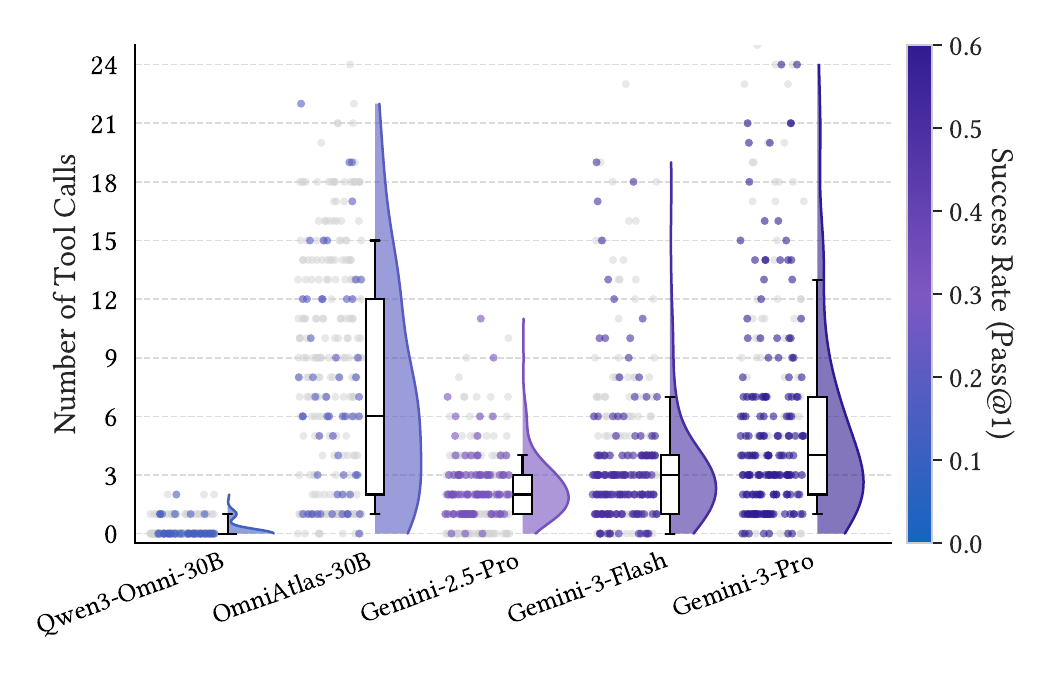}
\vspace{-1.35\baselineskip}
\caption{\textbf{Tool call distribution analysis.} Dots indicate individual task runs (grey: failed; colored: successful). The box and half-violin plots visualize the tool call frequency for successful runs.}
\label{fig:turn_distribution}
\vspace{-1.2\baselineskip}
\end{wrapfigure}

\Cref{fig:turn_distribution} illustrates the distribution of tool calls per task run, highlighting successful runs in color.

\textit{\textbf{(1) External tools are indispensable:}} Models exhibiting minimal tool usage (e.g., Qwen-3-Omni-30B, concentrated near 0 calls) achieve negligible success rates. This confirms that native perception alone is insufficient for many OmniGAIA tasks, necessitating external evidence gathering.

\textit{\textbf{(2) More tool calls do not guarantee better performance:}} A high volume of tool calls (long tails reaching $>10$--20) does not guarantee success. A substantial fraction of such runs still fail, indicative of inefficient exploration or ``thrashing'' behaviors where models repeatedly invoke tools without resolving underlying uncertainties.

\textit{\textbf{(3) OmniAtlas shifts from under-calling to more active tool use:}} In contrast to the passive Qwen-3-Omni-30B, OmniAtlas-30B exhibits a much higher and broader tool-call distribution, aligning with its improvements in ineffective tool-use and 
overall Pass@1, while leaving new opportunities for more efficient and effective tool-use policies.

\subsection{Native Perception vs.\ Tool-based Perception}
\begin{wraptable}{r}{0.52\linewidth}
\vspace{-0.9\baselineskip}
\centering
\caption{\textbf{Performance analysis with tool-based perception.} All Qwen-3 models use the 30B-A3B version for fair comparison. Best results within the Gemini-3 and Qwen-3 groups are in \textbf{bold}.}
\label{tab:tool_perception_analysis}
\setlength\tabcolsep{4.5pt}
\resizebox{\linewidth}{!}{%
\begin{tabular}{l c c c c c c}
\toprule
\multirow{2}[2]{*}{{\bfseries Method}} &
\multirow{2}[2]{*}{{\bfseries \makecell{Perception\\Model}}} &
\multicolumn{3}{c}{{\bfseries Difficulty Levels}} &
\multirow{2}[2]{*}{{\bfseries Avg.}} &
\multirow{2}[2]{*}{{\bfseries \makecell{Tool\\Calls}}} \\
\cmidrule(lr){3-5}
 & & Easy & Med. & Hard & & \\
\midrule
\multicolumn{7}{l}{\textit{{Native Omni-Modal Perception (Input All Media)}}} \\
Gemini-3-Flash & No Need & \cellcolor{geminiBlue!70}{\bfseries 67.2} & \cellcolor{geminiBlue!53}{\bfseries 46.9} & \cellcolor{geminiBlue!70}{\bfseries 37.2} & \cellcolor{geminiBlue!70}{\bfseries 51.7} & \cellcolor{geminiBlue!20}{\bfseries 4.4} \\
Qwen-3-Omni & No Need & \cellcolor{qwenBlue!26}19.7 & \cellcolor{qwenBlue!20}10.6 & \cellcolor{qwenBlue!70}{\bfseries 9.0} & \cellcolor{qwenBlue!29}13.3 & \cellcolor{qwenBlue!20}{\bfseries 0.2} \\
\midrule
\multicolumn{7}{l}{\textit{{Audio Perception Model as a Tool (Input Only Vision)}}} \\
Gemini-3-Flash & Gemini-3-Flash & \cellcolor{geminiBlue!51}60.7 & \cellcolor{geminiBlue!70}48.8 & \cellcolor{geminiBlue!53}35.9 & \cellcolor{geminiBlue!60}50.0 & \cellcolor{geminiBlue!52}7.6 \\
Qwen-3-Omni & Qwen-3-Omni & \cellcolor{qwenBlue!42}24.6 & \cellcolor{qwenBlue!49}15.0 & \cellcolor{qwenBlue!20}3.9 & \cellcolor{qwenBlue!51}15.8 & \cellcolor{qwenBlue!32}0.8 \\
Qwen-3-VL & Qwen-3-Omni & \cellcolor{qwenBlue!42}24.6 & \cellcolor{qwenBlue!70}{\bfseries 18.1} & \cellcolor{qwenBlue!57}7.7 & \cellcolor{qwenBlue!70}{\bfseries 18.1} & \cellcolor{qwenBlue!70}2.8 \\
\midrule
\multicolumn{7}{l}{\textit{{Visual Perception Model as a Tool (Input Only Audio)}}} \\
Gemini-3-Flash & Gemini-3-Flash & \cellcolor{geminiBlue!20}50.0 & \cellcolor{geminiBlue!20}43.1 & \cellcolor{geminiBlue!20}33.3 & \cellcolor{geminiBlue!20}43.3 & \cellcolor{geminiBlue!44}6.8 \\
Qwen-3-Omni & Qwen-3-Omni & \cellcolor{qwenBlue!20}18.0 & \cellcolor{qwenBlue!25}11.3 & \cellcolor{qwenBlue!32}5.1 & \cellcolor{qwenBlue!20}12.2 & \cellcolor{qwenBlue!26}0.5 \\
\midrule
\multicolumn{7}{l}{\textit{{Audio and Visual Perception Models as Tools (Input No Media)}}} \\
Gemini-3-Flash & Gemini-3-Flash & \cellcolor{geminiBlue!27}52.5 & \cellcolor{geminiBlue!53}46.9 & \cellcolor{geminiBlue!53}35.9 & \cellcolor{geminiBlue!39}46.4 & \cellcolor{geminiBlue!70}9.4 \\
Qwen-3-Omni & Qwen-3-Omni & \cellcolor{qwenBlue!40}23.8 & \cellcolor{qwenBlue!29}11.9 & \cellcolor{qwenBlue!57}7.7 & \cellcolor{qwenBlue!44}15.0 & \cellcolor{qwenBlue!55}2.0 \\
Qwen-3 & Qwen-3-Omni & \cellcolor{qwenBlue!70}{\bfseries 32.8} & \cellcolor{qwenBlue!20}10.6 & \cellcolor{qwenBlue!45}6.4 & \cellcolor{qwenBlue!62}17.2 & \cellcolor{qwenBlue!60}2.3 \\
\bottomrule
\end{tabular}}
\vspace{-1.2\baselineskip}
\end{wraptable}

Do we really need native omni-modal agents, or can perception tools substitute for them? \Cref{tab:tool_perception_analysis} offers a controlled ablation under matched model families.

\textit{\textbf{(1) Native perception is optimal for strong agents:}} For Gemini-3-Flash, native perception achieves the best Avg.\ score (51.7) with fewer tool calls (4.4). Replacing native channels with perception tools lowers Avg.\ to 50.0/43.3/46.4 while increasing calls to 7.6/6.8/9.4, yielding no accuracy-cost benefit.

\textit{\textbf{(2) Perception tools help weak agents on Easy and Medium but not Hard:}} For Qwen-3-Omni, tools improve Easy/Med.\ performance (19.7$\rightarrow$24.6; 10.6$\rightarrow$15.0/11.9) but consistently reduce Hard performance (9.0$\rightarrow$3.9/5.1/7.7). This suggests tool outputs can patch missing low-level signals, but cannot replace native cross-modal integration for long-horizon reasoning.

\textit{\textbf{(3) Tool perception consistently increases interaction cost:}} Adding perception tools increases the call budget across settings (Qwen-3-Omni: 0.2$\rightarrow$0.5--2.0; Gemini-3-Flash: 4.4$\rightarrow$6.8--9.4), implying higher latency and deployment cost.

Therefore, native perception should be the default for capable omni-modal agents to achieve higher performance ceilings, while tool-based perception is best treated as a fallback for weaker agents or missing-modality scenarios.

\newpage

\subsection{Training Effectiveness of OmniAtlas}
\begin{wraptable}{r}{0.52\linewidth}
\vspace{-1.12\baselineskip}
\centering
\caption{\textbf{Training effectiveness of OmniAtlas.} We report four primary error types ($\downarrow$) and the overall performance ($\uparrow$). For each model group, the best result in each column is highlighted in \textbf{bold}.}
\label{tab:omniatlas_training_effectiveness}
\setlength\tabcolsep{3pt}
\resizebox{\linewidth}{!}{%
\begin{tabular}{l c c c c c}
\toprule
{\bfseries Method} &
{\bfseries \makecell{Visual\\Percept.}} &
{\bfseries \makecell{Audio\\Percept.}} &
{\bfseries \makecell{Ineffect.\\Tool-Use}} &
{\bfseries \makecell{Reason.\\Error}} &
{\bfseries Perform.} \\
\midrule
{Qwen-2.5-Omni-7B} & \cellcolor{geminiBlue!70}41.4 & \cellcolor{geminiBlue!45}48.3 & \cellcolor{geminiBlue!70}91.9 & \cellcolor{geminiBlue!70}78.6 & \cellcolor{geminiBlue!20}3.6 \\
{~+ OmniAtlas-SFT} & \cellcolor{geminiBlue!45}38.9 & \cellcolor{geminiBlue!70}49.7 & \cellcolor{geminiBlue!45}69.2 & \cellcolor{geminiBlue!45}75.0 & \cellcolor{geminiBlue!45}11.4 \\
{~~~+ OmniDPO} & \cellcolor{geminiBlue!20}{\bfseries 37.2} & \cellcolor{geminiBlue!20}{\bfseries 46.1} & \cellcolor{geminiBlue!20}{\bfseries 67.2} & \cellcolor{geminiBlue!20}{\bfseries 72.8} & \cellcolor{geminiBlue!70}{\bfseries 13.3} \\
\midrule
{Qwen-3-Omni-30B} & \cellcolor{qwenBlue!45}31.7 & \cellcolor{qwenBlue!45}33.9 & \cellcolor{qwenBlue!70}81.1 & \cellcolor{qwenBlue!70}79.7 & \cellcolor{qwenBlue!20}13.3 \\
{~+ OmniAtlas-SFT} & \cellcolor{qwenBlue!70}32.2 & \cellcolor{qwenBlue!70}35.8 & \cellcolor{qwenBlue!45}65.3 & \cellcolor{qwenBlue!45}68.1 & \cellcolor{qwenBlue!45}18.9 \\
{~~~+ OmniDPO} & \cellcolor{qwenBlue!20}{\bfseries 30.3} & \cellcolor{qwenBlue!20}{\bfseries 31.9} & \cellcolor{qwenBlue!20}{\bfseries 59.4} & \cellcolor{qwenBlue!20}{\bfseries 64.4} & \cellcolor{qwenBlue!70}{\bfseries 20.8} \\
\bottomrule
\end{tabular}}
\vspace{-1.2\baselineskip}
\end{wraptable}
\Cref{tab:omniatlas_training_effectiveness} quantifies how OmniAtlas-SFT and OmniDPO affect error rates and performance. 

\textit{\textbf{(1) OmniAtlas-SFT contributes most of the gains:}} It drives the majority of improvements by boosting Pass@1 and reducing the ineffective tool-use rate (Qwen-3-Omni-30B: 13.3$\rightarrow$18.9, 81.1\%$\rightarrow$65.3\%). 

\textit{\textbf{(2) OmniDPO further delivers across-the-board gains:}} It provides additional improvements (to 13.3$\rightarrow$20.8) and continues to lower perception, tool-use, and reasoning errors, which verifies the effectiveness of the fine-grained error correction.

\section{Conclusion and Future Work}
We introduce \textit{\textbf{OmniGAIA}}, a benchmark for \emph{native omni-modal agents} that requires multi-hop reasoning and multi-turn tool use over video-with-audio and image+audio inputs. OmniGAIA is built with an event-graph pipeline that aligns and expands cross-modal evidence with tools, then synthesizes verifiable multi-hop questions via controllable event fuzzification and human-vetted screening. We further propose \textit{\textbf{OmniAtlas}}, a native omni-modal foundation agent that follows tool-integrated reasoning with \emph{active} perception, trained via hindsight-guided tree exploration, trajectory-level masked SFT, and \textit{OmniDPO} for fine-grained error correction. Experiments show OmniGAIA remains challenging for current models, and that effective tool-use and long-horizon reasoning---rather than parameter scaling alone---are decisive bottlenecks; our OmniAtlas recipe improves Qwen3-Omni from 13.3 to 20.8 Pass@1 while reducing tool-use and reasoning failures.

Looking ahead, we see three promising directions: \textit{\textbf{(1) Omni-modal Agentic RL}} to directly optimize long-horizon agentic policies under omni-modal feedback; \textit{\textbf{(2) Omni-modal MCP Services}}  with scalable tools for broader omni-modal tasks; and \textit{\textbf{(3) Omni-modal Embodied Agents}} benchmarks and foundation models in physical world, advancing LLM-brained AI assistants for real-world task completion.

\section{Impact Statement}
This work advances research on \emph{native omni-modal agents} by introducing OmniGAIA, a benchmark for long-horizon multi-hop reasoning with multi-turn tool use over video-with-audio and image+audio inputs, and by proposing OmniAtlas, a practical recipe for improving such tool-integrated behaviors in open models. These contributions may enable more reliable cross-modal grounding and verification in assistive applications (e.g., education and accessibility) and help standardize evaluation of tool-augmented omni-modal agents. We emphasize that any omni-modal agent should respect data provenance and licensing and prioritize privacy-preserving practices when handling audio/visual inputs.

\bibliographystyle{plainnat}
\bibliography{citation}

\clearpage
\appendix

\section*{Appendix}


\section{Implementation Details}
\label{app:implementation_details}

\tcbset{
  omnibox/.style={
    enhanced jigsaw,
    breakable,
    pad at break*=1mm,
    colframe=gray,
    colback=gray!5!white,
    coltitle=white,
    coltext=black,
    fonttitle=\bfseries,
    boxrule=1pt,
    arc=2mm,
    width=\linewidth,
    left=7pt,
    right=7pt,
    top=4pt,
    bottom=4pt
  },
  omniboxlisting/.style={
    omnibox,
    listing only,
    listing options={
      basicstyle=\ttfamily\fontsize{7.2pt}{8.8pt}\selectfont,
      breaklines=true,
      breakatwhitespace=false,
      columns=fullflexible,
      keepspaces=true
    }
  }
}

\subsection{Training Details}
\label{app:training_details}

We implement omni-modal agentic SFT and DPO training based on the LlamaFactory codebase~\citep{LlamaFactory}. Following Section~\ref{sec:OmniAtlas}, we first perform supervised fine-tuning for 2 epochs on 2{,}156 synthesized high-quality trajectories, and then continue training with OmniDPO for another 2 epochs to obtain the final OmniAtlas models. We train three backbone scales: Qwen2.5-Omni-[3B,7B]~\citep{Qwen2.5-Omni}, and Qwen3-Omni-30B-A3B-Thinking~\citep{Qwen3-Omni}. All model parameters are updated during training, including the vision tower, multi-modal projector, and language model. The training experiments were conducted on four nodes of 8 NVIDIA H20-141GB GPUs.

\subsection{Evaluation Details}
\label{app:evaluation_details}

We evaluate models using a two-stage procedure that combines exact match with an LLM-as-a-Judge fallback.
Given a question, we first attempt to extract the model predicted answer enclosed by \texttt{<answer>} and \texttt{</answer>}. If an extracted answer exists, we perform an exact string match against the labeled answer. If it matches exactly, the prediction is marked as correct and no LLM judging is used.
If an extracted answer exists but does not exactly match, or if no \texttt{<answer>...</answer>} span can be extracted, we ignore the extracted span (if any), take the last 20 words of the model output (split by spaces) as the predicted answer, and use LLM-as-a-Judge to determine whether it is equivalent to the labeled answer.

\section{Instruction Templates}
\label{app:instructions}

\subsection{Evaluation Prompt}
This prompt implements our LLM-as-a-Judge step for answer equivalence when exact match is insufficient.
We constrain the judge to output a single binary label (\texttt{Correct}/\texttt{Incorrect}) to make Pass@1 computation deterministic and avoid leaking intermediate reasoning. The judge takes only the question, the normalized prediction string (the last 20 words), and the labeled answer.
\begin{tcblisting}{omniboxlisting,title=LLM-as-a-Judge Prompt (DeepSeek-V3.2)}
Please determine if the model correctly predicted the answer.
Question: {question}
Model Predicted Answer: {predicted}
Labeled Answer: {standard}
Return 'Correct' if the model's prediction is completely accurate, otherwise return 'Incorrect'. Provide only this single word response.
\end{tcblisting}

\subsection{System Prompts}
\label{app:system_prompts}

\vspace{0.1cm}
This is the unified system prompt used for all base agents in our evaluation to standardize instruction-following, tool usage, and answer formatting across models.
We explicitly require the final answer to be wrapped by \texttt{<answer>...</answer>} so it can be reliably extracted for exact matching and judging, while leaving the model free to use tools and multi-step reasoning internally.
\begin{tcblisting}{omniboxlisting,title=Base Agent System Prompt}
You are an omni-modal general AI assistant. Please answer the question provided to you based on the input image, audio, or video content.

You should think step by step to answer the question. You may use available tools to assist with your analysis if needed.

Please provide your final answer using this format: <answer>YOUR_ANSWER</answer>.
\end{tcblisting}

\vspace{0.1cm}
This prompt equips OmniAtlas with \emph{active omni-modal perception}: when the model is uncertain about specific regions/segments, it can explicitly request additional evidence by calling perception tools.
The added note encourages ``look/listen-where-needed'' behavior (Section~\ref{sec:OmniAtlas}) instead of passively relying on a single lossy media ingestion, which is critical for long videos and high-resolution images.
\begin{tcblisting}{omniboxlisting,title=OmniAtlas System Prompt}
You are an omni-modal general AI assistant. Please answer the question provided to you based on the input image, audio, or video content.

You should think step by step to answer the question. You may use available tools to assist with your analysis if needed.

**Note:**
- If there are segments in the input image/audio/video that are unclear to you, you should use the "read_image/read_audio/read_video" tool to examine them carefully to ensure you have correctly perceived the input media.

Please provide your final answer using this format: <answer>YOUR_ANSWER</answer>.
\end{tcblisting}

\subsection{Active Omni-Modal Perception Tool Schemas}
\label{app:active_perception_tools}

\vspace{0.1cm}
This schema defines the \texttt{read\_video} tool used by OmniAtlas to retrieve a specific time window from a long video for higher-fidelity inspection.
Exposing \texttt{t\_start}/\texttt{t\_end} enables targeted evidence acquisition and reduces unnecessary context/cost versus loading the entire video.
\begin{tcblisting}{omniboxlisting,title=Function Schema: read\_video}
def get_function_schema_read_video():
    return {
        "type": "function",
        "function": {
            "name": "read_video",
            "description": "Reads a specific time segment of a video file to examine details.",
            "parameters": {
                "type": "object",
                "properties": {
                    "video_id": {"type": "string", "description": "The video identifier or filename."},
                    "t_start": {"type": "integer", "description": "Start time in seconds."},
                    "t_end": {"type": "integer", "description": "End time in seconds."},
                },
                "required": ["video_id", "t_start", "t_end"],
            },
        },
    }
\end{tcblisting}

\vspace{0.1cm}
This schema defines the \texttt{read\_audio} tool for selectively listening to a specific time segment.
Segment-level access supports pinpointing key speech/non-speech cues and mitigates information loss from global summaries.
\begin{tcblisting}{omniboxlisting,title=Function Schema: read\_audio}

def get_function_schema_read_audio():
    return {
        "type": "function",
        "function": {
            "name": "read_audio",
            "description": "Reads a specific time segment of an audio file to listen to details.",
            "parameters": {
                "type": "object",
                "properties": {
                    "audio_id": {"type": "string", "description": "The audio identifier or filename."},
                    "t_start": {"type": "integer", "description": "Start time in seconds."},
                    "t_end": {"type": "integer", "description": "End time in seconds."},
                },
                "required": ["audio_id", "t_start", "t_end"],
            },
        },
    }
\end{tcblisting}

\vspace{0.1cm}
This schema defines the \texttt{read\_image} tool to re-examine images, optionally with a crop box.
Cropping enables fine-grained verification of small objects/text without downsampling the entire image, aligning with our active perception principle.
\begin{tcblisting}{omniboxlisting,title=Function Schema: read\_image}

def get_function_schema_read_image():
    return {
        "type": "function",
        "function": {
            "name": "read_image",
            "description": "Reads specific images to view them in detail. Optionally crop the image by providing a crop box [left, top, right, bottom].",
            "parameters": {
                "type": "object",
                "properties": {
                    "image_ids": {"type": "array", "items": {"type": "string"}, "description": "List of image identifiers or filenames."},
                    "crop_box": {
                        "type": "array",
                        "items": {"type": "integer"},
                        "minItems": 4,
                        "maxItems": 4,
                        "description": "Optional. A 4-element list [left, top, right, bottom] specifying the cropping rectangle.",
                    },
                },
                "required": ["image_ids"],
            },
        },
    }
\end{tcblisting}

\subsection{Tool-based Perception: Tool Schemas and System Prompts}
\label{app:tool_based_perception}

\vspace{0.1cm}
This schema defines the \texttt{audio\_qa} perception tool used in our tool-based perception ablations (Table~\ref{tab:tool_perception_analysis}) to answer sub-questions from audio only.
By wrapping a perception model behind a tool interface, we can isolate whether deficiencies come from perception versus agentic planning/tool use.
\begin{tcblisting}{omniboxlisting,title=Perception Tool Schema: audio\_qa}
def get_openai_function_audio_qa() -> Dict[str, Any]:
    return {
        "type": "function",
        "function": {
            "name": "audio_qa",
            "description": "Answer the question using audio from audio_path or video_path.",
            "parameters": {
                "type": "object",
                "properties": {
                    "question": {"type": "string", "description": "The question to answer."},
                    "audio_path": {"type": "string", "description": "Audio file path."},
                    "video_path": {"type": "string", "description": "Video file path (audio will be used)."},
                },
                "required": ["question"],
            },
        },
    }
\end{tcblisting}

\vspace{0.1cm}
This schema defines the \texttt{vision\_qa} perception tool for answering sub-questions from visual content only.
Together with \texttt{audio\_qa}, it enables controlled settings where the base agent can delegate missing modalities to specialized tools.
\begin{tcblisting}{omniboxlisting,title=Perception Tool Schema: vision\_qa}

def get_openai_function_vision_qa() -> Dict[str, Any]:
    return {
        "type": "function",
        "function": {
            "name": "vision_qa",
            "description": "Answer the question using visual content from image_path or video_path.",
            "parameters": {
                "type": "object",
                "properties": {
                    "question": {"type": "string", "description": "The question to answer."},
                    "image_path": {"type": "string", "description": "Image file path."},
                    "video_path": {"type": "string", "description": "Video file path."},
                },
                "required": ["question"],
            },
        },
    }
\end{tcblisting}

\vspace{0.1cm}
This is the system prompt for the \texttt{audio\_qa} tool backend.
The prompt strictly restricts the tool to audio evidence and allows abstention (``cannot determine'') to prevent hallucinated cross-modal guesses.
\begin{tcblisting}{omniboxlisting,title=System Prompt: Audio QA Prompt}
You are an audio perception assistant. Answer the question using only the provided audio. If the audio does not contain enough information, say you cannot determine.
\end{tcblisting}

\vspace{0.1cm}
This is the system prompt for the \texttt{vision\_qa} tool backend.
It enforces a vision-only evidence policy and abstention when visual information is insufficient, making the ablation faithful and verifiable.
\begin{tcblisting}{omniboxlisting,title=System Prompt: Vision QA Prompt}
You are a visual perception assistant. Answer the question using only the provided image or video. If the visual content does not contain enough information, say you cannot determine.
\end{tcblisting}

\subsection{Perception Analysis Prompts for Data Construction}
\label{app:perception_analysis_prompts}

\vspace{0.1cm}
This prompt converts raw images into a high-density, structured JSON report (OCR, objects, faces, global summary) used as intermediate signals for event graph construction.
The ``certainty-first'' constraint reduces noise and hallucination in downstream graph reasoning, while the structured fields make evidence retrieval and linking explicit.
\begin{tcblisting}{omniboxlisting,title=Image Analysis Prompt}
Please analyze this image and provide a comprehensive structured report in strictly JSON format.

**Crucial Guidelines:**
1. **CERTAINTY FIRST**: Only provide information you are absolutely certain about. Do not guess or hallucinate details from blurry or ambiguous regions. If you are unsure, do not include it.
2. **COMPREHENSIVE & FACTUAL**: Focus on extracting strictly factual, objective information. Ensure high detail and density of information to support future analysis. Describe all visible text, objects, and people in detail.

The JSON object must contain the following fields:

```json
{
    "ocr": [
        { "text": "detected text string", "detailed_features": "detailed features of the text (e.g. location, color, etc)" }
    ],
    "objects": [
        { "label": "object name", "confidence": 0.95, "detailed_features": "detailed features of the object (e.g. location, color, shape, texture, etc)" }
    ],
    "faces": [
        {
            "age": "estimated age range (e.g. 25-30)",
            "gender": "Male/Female",
            "expression": "detailed description of facial expression and emotion",
            "visual_attributes": "clothing, glasses, hair color, distinctive features",
            "activity": "what the person is specifically doing (e.g. reading a book, talking on phone, typing on laptop, cooking, exercising, etc)"
        }
    ],
    "global_summary": "A comprehensive, exhaustive, and highly detailed description of the image content, covering all visible elements, context, actions, and details."
}
```

If a field is not applicable or nothing is detected, return an empty list for that field.
After thinking, output your final response as a JSON code block:

```json
{...}
```
\end{tcblisting}

\vspace{0.1cm}
This clip-level prompt produces fine-grained audio annotations for a specific time window, enabling time-aligned evidence mining for long recordings.
Clip segmentation improves recall of transient cues (short utterances/events) and supports our pipeline's timestamped linking and multi-hop reasoning.
\begin{tcblisting}{omniboxlisting,title=Audio Clip Analysis Prompt}
You are analyzing a specific audio clip segment from a longer audio/video.
**Clip Context:**
- Total Duration: {total_duration:.2f} seconds
- Current Clip Range: {start_time:.2f}s to {end_time:.2f}s

Please analyze this short audio clip and provide a comprehensive structured report in strictly JSON format.

**Crucial Guidelines:**
1. **CERTAINTY FIRST**: Only provide information you are absolutely certain about. Do not guess unclear speech or ambiguous sounds. If you are unsure, do not include it.
2. **COMPREHENSIVE & FACTUAL**: Concentrate on factual information from both speech and non-speech sounds. Ensure high detail and density of information to support future analysis.

The JSON object must contain the following fields:

```json
{
    "asr": [
        { "text": "transcribed text", "start": 0.0, "end": 2.5, "speaker": "speaker_1" }
    ],
    "speakers": {
        "speaker_1": {
            "gender": "Male/Female",
            "age_estimate": "Adult/Child/Elderly",
            "accent": "description of accent or dialect",
            "tone": "emotional tone (e.g. anxious, authoritative, calm)"
        }
    },
    "events": [
        {
            "label": "event name (e.g. dog barking, siren, applause)",
            "category": "environment/sound_effect/music/speech",
            "start": 1.2,
            "end": 3.5
        }
    ],
    "nonspeech_information": "detailed description of the non-speech information in this clip, focusing on factual and certain information",
    "global_summary": "A comprehensive, exhaustive, and highly detailed summary of the clip content, covering all speech, sound events, background noises, and emotional cues, focusing on factual and certain information"
}
```

If no speech is detected, 'asr' should be an empty list.
If no specific events are detected, 'events' should be an empty list.
After thinking, output your final response as a JSON code block:

```json
{...}
```
\end{tcblisting}

\vspace{0.1cm}
This global prompt produces an overall structured audio report for the entire recording, complementing clip-level details with global context.
We keep the same schema as the clip prompt to make aggregation consistent, while allowing longer ASR segmentation and higher-level summaries for event graph nodes.
\begin{tcblisting}{omniboxlisting,title=Audio Global Analysis Prompt}
Please analyze this audio and provide a comprehensive structured report in strictly JSON format.

**Crucial Guidelines:**
1. **CERTAINTY FIRST**: Only provide information you are absolutely certain about. Do not guess unclear speech or ambiguous sounds. If you are unsure, do not include it.
2. **COMPREHENSIVE & FACTUAL**: Concentrate on factual information from both speech and non-speech sounds. Ensure high detail and density of information to support future analysis.

The JSON object must contain the following fields:

```json
{
    "asr": [
        // If a single person speaks for a long time, segment the speech into pieces, with each segment containing one piece of information
        { "text": "transcribed text", "start": 0.0, "end": 2.5, "speaker": "speaker_1" }
    ],
    "speakers": {
        "speaker_1": {
            "gender": "Male/Female",
            "age_estimate": "Adult/Child/Elderly",
            "accent": "description of accent or dialect",
            "tone": "emotional tone (e.g. anxious, authoritative, calm)"
        }
    },
    "events": [
        {
            "label": "event name (e.g. dog barking, siren, applause)",
            "category": "environment/sound_effect/music/speech",
            "start": 1.2,
            "end": 3.5
        }
    ],
    "nonspeech_information": "detailed description of the non-speech information in the audio, focusing on factual and certain information",
    "global_summary": "A comprehensive, exhaustive, and highly detailed summary of the audio content, covering all speech, sound events, background noises, and emotional cues, focusing on factual and certain information"
}
```

If no speech is detected, 'asr' should be an empty list.
If no specific events are detected, 'events' should be an empty list.
After thinking, output your final response as a JSON code block:

```json
{...}
```
\end{tcblisting}

\subsection{Error Analysis Prompt}
\label{app:error_analysis_prompt}

\vspace{0.1cm}
This prompt supports our fine-grained error taxonomy analysis (Figure~\ref{fig:error_analysis_heatmap}) by labeling failure causes from full execution traces.
We allow multi-label categorization to capture cascade failures (e.g., tool misuse leading to reasoning errors) and require a JSON output for easy aggregation and reproducibility.
\begin{tcblisting}{omniboxlisting,title=Fine-grained Error Analysis Prompt}
You are an expert AI system analyst. The following is a trace of an AI agent attempting to solve a multimodal question but failing.

**Task Information**:
Question: {question}
Omni Modal Input: {omni_modal_input}
Annotated Solution: {annotated_solution}
Correct Answer: {answer}

**Agent Execution Trace**:
{trace_str}

**Analysis Request**:
Identify all the causes of the error. You may select MULTIPLE categories from the list below if applicable.

Visual Perception Error: Agent misread or failed to identify specific visual details from media.
Audio Perception Error: Agent misheard or failed to identify specific audio details from media.
Ineffective Tool Call: Agent failed to make necessary tool calls (such as web search, page browsing, or code execution), or used tools but did not obtain the required information.
Reasoning Error: Agent found all correct facts but made logical mistakes or invalid assumptions.
Instruction Following Error: Agent misunderstood the question or failed to follow constraints.
No Answer: Agent failed to provide a final answer.

**Output Format**:
Return the analysis in a valid JSON object wrapped in markdown code blocks. The "categories" field must be a list of strings matching the category names above.

```json
{
  "categories": ["...", ...],
  "explanation": "..."
}
```
\end{tcblisting}

\section{Detailed Related Work}

\subsection{Omni-Modal Foundation Models and Benchmarks}
Building on advances in pure-text~\citep{llama3}, vision-language~\citep{GPT-4o}, and audio-language~\citep{Qwen2-Audio} foundation models, recent omni-modal models seek to unify text, vision, and audio within a single LLM backbone. A common approach adopts a unified tokenization-and-projection interface that maps heterogeneous visual and acoustic inputs into a shared token space~\citep{Qwen3-Omni,NEXUS-O,NExT-OMNI,OmniVinci,JavisGPT}. Concurrent work further strengthens omni-modal reasoning behaviors~\citep{Omni-R1,OmnixR,ELLSA,OmniVideo-R1}, token compression~\citep{OmniSIFT}, and reward modeling~\citep{Omni-Reward,Omni-RRM}.
For evaluation, existing benchmarks (e.g., OmniBench~\citep{OmniBench}, WorldSense~\citep{WorldSense} and Daily-Omni~\citep{Daily-Omni}) largely emphasize short audios/videos and perception-centric tasks, leaving long-horizon reasoning and tool-integrated agency underexplored. This gap hinders complex, interactive real-world applications.

\subsection{Autonomous Agents}
LLM-driven autonomous agents tackle real-world tasks by reasoning and acting through external tools that interface with their environment~\citep{wang_llm_agent_survey,LLM-Agent-Survey,GISA,MCPMark}. 
Existing approaches broadly fall into workflow-based paradigms~\citep{yao2022react,Plan-and-Solve,HiAgent} and native agentic reasoning methods~\citep{2501_search_o1, ToRL,WebThinker,ReTool,VerlTool,Tool-Star,WebDancer,HiRA,AEPO,DeepAgent,ET-Agent,hu2026memory,HierSearch,Dr.Zero}, and have shown strong performance on text-only tasks. Moving beyond text, recent studies investigate vision-language agents for multimodal web search~\citep{MM-BrowseComp,MMSearch-R1,WebWatcher}, long-form video understanding~\citep{VideoAgent,VideoExplorer,Deep-Video-Discovery,VideoARM,Agent-Omni,OmniAgent,OmniRAG-Agent}, and GUI navigation~\citep{OSWorld,AppAgent,Mobile-Agent,PhoStream}. However, \emph{omni-modal} foundation agents that natively fuse audio, vision, and language while performing long-horizon agentic reasoning remain underexplored. Such capabilities are essential for building general AI assistants in real-world scenarios.

\begin{table*}[!t]
\centering
\caption{\textbf{Case Study I (Failure).} Qwen3-Omni-30B-A3B on an OmniGAIA video-with-audio question. The model fails to ground the bridge to the Joliet Iron Works context and does \emph{not} invoke any external tools (0 tool calls), leading to an unverified Chicago-bridge prior and an incorrect final answer. (\texttt{[...]} denotes omitted trace content for brevity.)}
\fontsize{9pt}{11pt}\selectfont
\begin{tabular}{p{0.98\linewidth}}
\midrule
\rowcolor{gray!20}{\bfseries Example from OmniGAIA} \\
\midrule
{\bfseries Question:} \\
During a visit to the Joliet Iron Works Historic Site as shown in the video, the speaker spots a movable bridge in the distance and remarks that it reminds him of a bridge featured in the movie \textit{The Blues Brothers}. What is the name of this bridge, and how many years had it been standing when filming for \textit{The Blues Brothers} began?\\
{\bfseries Labeled Answer:} \\
Ruby Street Bridge; 44 \\
\midrule
\rowcolor{gray!20}
{\bfseries Model Output by Qwen3-Omni-30B-A3B}\\
\midrule
{\bfseries Tool Calls:} 0 (tools available but unused).\\
(\textit{Condensed}) The model notes a potential mismatch between the question context ({\bfseries Joliet Iron Works}, the speaker pointing to ``a bridge \dots it's going down'') and its movie-location prior (well-known Chicago bridges in \textit{The Blues Brothers}). It briefly considers whether Joliet landmarks (e.g., Old Joliet Prison) could imply a local bridge, but ultimately treats the question as asking for the bridge \emph{featured in the movie} and selects the {\bfseries LaSalle Street Bridge} in Chicago. Without using tools to verify, it assumes a completion year of 1928, takes filming to begin in 1979, and computes \(1979 - 1928 = 51\) years, yielding \texttt{<answer>LaSalle Street Bridge, 51</answer>}. \texttt{[...]}\\
\midrule
\rowcolor{gray!20}{\bfseries Error Analysis}\\
\midrule
{\bfseries Error Categories:} Visual Perception Error; Ineffective Tool Call; Reasoning Error.\\
{\bfseries Why it fails:} (i) It does not use the video/audio context to anchor the bridge to Joliet; (ii) it under-calls tools despite tool availability; (iii) it relies on an unverified prior about a Chicago bridge and its construction year, producing an incorrect age.\\
{\bfseries Evaluation:} EM=0, LLM-as-a-Judge=\texttt{Incorrect}.\\
\bottomrule
\end{tabular}
\label{tab:case_video_qa}
\end{table*}

\begin{table*}[!t]
\centering
\caption{\textbf{Case Study II (Failure).} OmniAtlas-Qwen3-30B-A3B on the same OmniGAIA instance. The model calls tools but exhibits query drift and confirmation bias, resulting in correct computation over incorrect facts. (\texttt{[...]} denotes omitted trace content for brevity.)}
\fontsize{9pt}{11pt}\selectfont
\begin{tabular}{p{0.98\linewidth}}
\midrule
\rowcolor{gray!20}{\bfseries Example from OmniGAIA} \\
\midrule
{\bfseries Question:} \\
During a visit to the Joliet Iron Works Historic Site as shown in the video, the speaker spots a movable bridge in the distance and remarks that it reminds him of a bridge featured in the movie *The Blues Brothers*. What is the name of this bridge, and how many years had it been standing when filming for The Blues Brothers commenced in July 1979?\\
{\bfseries Labeled Answer:} \\
Ruby Street Bridge; 44 \\
\midrule
\rowcolor{gray!20}
{\bfseries Model Output by OmniAtlas-Qwen3-30B-A3B}\\
\midrule
I can see in the video that the speaker is at the Joliet Iron Works Historic Site, and I can hear him talking about the surroundings. He mentions seeing a bridge in the distance that reminds him of one from The Blues Brothers movie. Based on what I hear, the bridge is the LaSalle Street Bridge in Chicago. However, I need to verify this and calculate how many years it had been standing when filming for The Blues Brothers started in July 1979. Let me search for more information about this bridge.

\blue{<tool\_call>}
{"name": "web\_search", "arguments": {"query": "LaSalle Street Bridge Chicago The Blues Brothers filming July 1979 construction date"}}
\blue{</tool\_call>}

\purple{<tool\_response>}
[...] The Blues Brothers - Filming Locations of Chicago and Los Angeles [...] LaSalle Street Bridge, Chicago [...]
\purple{</tool\_response>}

Let me get more specific information about the LaSalle Street Bridge and its construction date.

\blue{<tool\_call>}
{"name": "web\_search", "arguments": {"query": "LaSalle Street Bridge construction date 1885 Chicago"}}
\blue{</tool\_call>}

\purple{<tool\_response>}
[...] Originally built as the Water Street Bridge in 1885, it was later renamed the LaSalle Street Bridge [...]
\purple{</tool\_response>}

Now I need to calculate how many years the LaSalle Street Bridge had been standing when filming for The Blues Brothers commenced in July 1979. According to the search results, the bridge was originally built in 1885. Let me calculate this:

\blue{<tool\_call>}
{"name": "code\_executor", "arguments": {"code": "\# Calculate years from 1885 to July 1979\textbackslash{}nconstruction\_year = 1885\textbackslash{}nfilming\_year = 1979\textbackslash{}nyears\_standing = filming\_year - construction\_year\textbackslash{}nprint(f'The LaSalle Street Bridge was built in {construction\_year}')\textbackslash{}nprint(f'Filming for The Blues Brothers commenced in July {filming\_year}')\textbackslash{}nprint(f'Number of years standing: {years\_standing}')"}}
\blue{</tool\_call>}

\purple{<tool\_response>}
The LaSalle Street Bridge was built in 1885
Filming for The Blues Brothers commenced in July 1979
Number of years standing: 94
\purple{</tool\_response>}

Let me verify the construction year with another source.

\blue{<tool\_call>}
{"name": "web\_search", "arguments": {"query": "\"Water Street Bridge\" 1885 LaSalle Street Bridge Chicago"}}
\blue{</tool\_call>}

\purple{<tool\_response>}
[...] the Water Street Bridge in 1885, it was later renamed the LaSalle Street Bridge after the street name was changed [...]
\purple{</tool\_response>}

Now I have clear information. Let me also search for more specific information about the filming dates for The Blues Brothers in Chicago.

\blue{<tool\_call>}
{"name": "web\_search", "arguments": {"query": "\"The Blues Brothers\" filming began July 1979 Chicago"}}
\blue{</tool\_call>}

\purple{<tool\_response>}
[...] Much of The Blues Brothers was shot on location in and around Chicago between July and October 1979 [...]
\purple{</tool\_response>}

Based on all the information gathered, I can now answer the question. Let me analyze what I've learned:
1. From the video: The speaker spots a movable bridge in the distance at the Joliet Iron Works Historic Site.
2. From the search results: The bridge referenced is the LaSalle Street Bridge (originally known as the Water Street Bridge) in Chicago.
3. From multiple sources: The bridge was constructed in 1885.
4. From the search results: Principal photography began in July 1979.
Therefore, the bridge spotted by the speaker at the Joliet Iron Works Historic Site is the LaSalle Street Bridge (originally known as the Water Street Bridge) in Chicago. It was built in 1885. When filming for The Blues Brothers commenced in July 1979, the bridge had been standing for 94 years (1979 - 1885 = 94 years).\\
\midrule
\rowcolor{gray!20}
{\bfseries Error Analysis}\\
\midrule
{\bfseries Main Errors:}
1. {\bfseries Incorrect Bridge Identification:} Model incorrectly identified the bridge as the {\bfseries LaSalle Street Bridge} in Chicago, while the correct answer is the {\bfseries Ruby Street Bridge} in Joliet (near the Joliet Iron Works Historic Site).
2. {\bfseries Incorrect Construction Year:} Model used 1885 as construction year, but Ruby Street Bridge was actually built in 1935.
3. {\bfseries Incorrect Age Calculation:} Based on wrong construction year, calculated 94 years instead of correct 44 years (1979 - 1935).

{\bfseries Reason for Errors:}
1. Over-relied on general knowledge about \textit{The Blues Brothers} movie (which features Chicago bridges) rather than focusing on the specific location mentioned in the question.
2. Failed to search for the specific bridge near Joliet Iron Works Historic Site, using instead generic Chicago bridge queries.
3. Did not verify local information about Joliet area bridges.

{\bfseries Correct Information:}
• {\bfseries Bridge Name:} Ruby Street Bridge
• {\bfseries Location:} Near Joliet Iron Works Historic Site, over Des Plaines River
• {\bfseries Construction Year:} 1935
• {\bfseries Filming Start:} July 1979
• {\bfseries Age at Filming:} 44 years (1979 - 1935)\\
\bottomrule
\end{tabular}
\label{tab:case_omniatlas_video_qa}
\end{table*}

\begin{table*}[!t]
\centering
\caption{\textbf{Case Study III (Success).} Gemini-3-Pro on the same OmniGAIA instance (condensed). The model grounds the location, retrieves the bridge construction year and filming start date with web search, uses code execution for arithmetic, and produces an answer judged equivalent to the label. (\texttt{[...]} denotes omitted trace content for brevity.)}
\fontsize{8.2pt}{10.2pt}\selectfont
\begin{tabular}{p{0.98\linewidth}}
\midrule
\rowcolor{gray!20}{\bfseries Example from OmniGAIA} \\
\midrule
{\bfseries Question:} \\
During a visit to the Joliet Iron Works Historic Site as shown in the video, the speaker spots a movable bridge in the distance and remarks that it reminds him of a bridge featured in the movie \textit{The Blues Brothers}. What is the name of this bridge, and how many years had it been standing when filming for \textit{The Blues Brothers} commenced in July 1979?\\
{\bfseries Labeled Answer:} \\
Ruby Street Bridge; 44 \\
\midrule
\rowcolor{gray!20}
{\bfseries Model Output by Gemini-3-Pro}\\
\midrule
The bridge is likely the Ruby Street Bridge in Joliet. I will verify the bridge name and its construction year, then compute its age in July 1979.

\blue{<tool\_call>}
{"name": "web\_search", "arguments": {"query": "Joliet Iron Works Historic Site Ruby Street Bridge Columbia Street east of Ruby Street Bridge"}}
\blue{</tool\_call>}

\purple{<tool\_response>}
[...] Joliet Iron Works Historic Site is located on Columbia Street, just east of the Ruby Street Bridge [...]
\purple{</tool\_response>}

\blue{<tool\_call>}
{"name": "web\_search", "arguments": {"query": "Ruby Street Bridge Joliet year built 1935"}}
\blue{</tool\_call>}

\purple{<tool\_response>}
[...] Ruby Street Bridge [...] Year Built: 1935 [...]
\purple{</tool\_response>}

\blue{<tool\_call>}
{"name": "web\_search", "arguments": {"query": "The Blues Brothers filming began July 1979"}}
\blue{</tool\_call>}

\purple{<tool\_response>}
[...] principal photography started in July 1979 [...]
\purple{</tool\_response>}

\blue{<tool\_call>}
{"name": "code\_executor", "arguments": {"code": "1979 - 1935"}}
\blue{</tool\_call>}

\purple{<tool\_response>}
44
\purple{</tool\_response>}

Based on the video and the location of the {\bfseries Joliet Iron Works Historic Site}, the movable bridge visible in the distance is the {\bfseries Ruby Street Bridge} (also known as the {\bfseries Ruby Street Bascule Bridge}, over the Des Plaines River).
\par\smallskip
The speaker remarks that the bridge reminds him of a bridge featured in \textit{The Blues Brothers}. While the famous bridge jump scene in the movie was filmed at the {\bfseries East 95th Street Bridge} in Chicago, the Ruby Street Bridge in Joliet has a similar appearance (both are bascule bridges) and is located near the Joliet Correctional Center (Joliet Prison), where the movie's opening scenes were filmed.
\par\smallskip
The Ruby Street Bridge was opened in {\bfseries 1935}. Filming for \textit{The Blues Brothers} began in {\bfseries July 1979}. Therefore, the bridge had been standing for {\bfseries 44} years when filming commenced.
\par\smallskip
\texttt{\textless answer\textgreater The bridge is the Ruby Street Bridge (or Ruby Street Bascule Bridge). It had been standing for 44 years when filming for The Blues Brothers commenced (1979 - 1935 = 44). \textless/answer\textgreater}\\
\midrule
\rowcolor{gray!20}
{\bfseries Evaluation}\\
\midrule
Exact match fails due to formatting differences, but LLM-as-a-Judge determines equivalence (\texttt{Correct}).\\
\bottomrule
\end{tabular}
\label{tab:case_gemini3pro_omnigaia}
\end{table*}

\section{Case Study}
\label{app:case_study}

We analyze three execution traces on the \emph{same} OmniGAIA instance (Tables~\ref{tab:case_video_qa}, \ref{tab:case_omniatlas_video_qa}, and~\ref{tab:case_gemini3pro_omnigaia}) to highlight a key lesson for omni-modal agents: {tool access is necessary but not sufficient}. The instance contains a deliberate distraction---the mention of \textit{The Blues Brothers}---which can trigger a strong \emph{Chicago-bridge prior}. The correct solution instead requires {location-first grounding} at {Joliet Iron Works} and then evidence-backed identification of the nearby movable bridge ({Ruby Street Bridge}, built 1935), followed by a simple computation for filming start in July 1979 (\(1979-1935=44\)).

\subsection{What Capabilities Does This Instance Stress?}
\label{app:case_required_capabilities}
This instance stresses a tightly-coupled chain of capabilities:
\begin{itemize}[leftmargin=*,nosep]
\item \textbf{Omni-modal grounding (location-first)}: anchor the bridge to the Joliet Iron Works context, instead of following movie-location priors.
\item \textbf{Tool planning \& query formulation}: issue \emph{entity- and location-specific} queries (e.g., ``Joliet Iron Works'' + ``Ruby Street Bridge''), rather than underspecified Chicago-centric searches.
\item \textbf{Hypothesis testing \& verification}: treat early guesses as hypotheses, and actively seek disconfirming/local evidence before committing to a bridge identity and construction year.
\item \textbf{Computation after verification}: use a calculator/code tool only after the facts are grounded (here, \(1979-1935\)).
\item \textbf{Answer normalization}: output a concise, extractable final answer aligned with the evaluation protocol.
\end{itemize}

\subsection{Case I: Failure by Under-Calling (No Tools)}
\label{app:case_failure_analysis_qwen3}
In Case I (Table~\ref{tab:case_video_qa}), the model fails early due to {premature closure} on a movie-driven prior. It does not use tools at all, so it never retrieves the decisive local evidence that ties the scene to the {Ruby Street Bridge} near Joliet Iron Works, nor does it verify the construction year and filming start date. As a result, it outputs a confident but unverified bridge name and an incorrect age.

\subsection{Case II: Failure by Tool-Query Drift (Tools Used, Wrong Hypothesis)}
\label{app:case_failure_analysis_omniatlas}
Case II (Table~\ref{tab:case_omniatlas_video_qa}) shows a different and more subtle failure: the model \emph{does} call tools, but its retrieval is locked onto the initial wrong hypothesis (a Chicago bridge). This produces confirmation bias: each search result reinforces the Chicago interpretation, and the agent never executes a corrective, location-specific query that could recover the Joliet bridge identity. The downstream computation is then correct arithmetic over wrong facts (e.g., subtracting from an incorrect construction year), illustrating that computation tools cannot compensate for upstream grounding/retrieval errors.

\subsection{Case III: Success via Location-First Disambiguation and Verification}
\label{app:case_success_analysis}
In Case III (Table~\ref{tab:case_gemini3pro_omnigaia}), the agent follows a robust pattern: it first {disambiguates by location} (Joliet Iron Works $\rightarrow$ nearby movable bridge), then performs {targeted retrieval} to verify the bridge identity and construction year (Ruby Street Bridge, 1935) and the filming start time (July 1979), and finally uses a computation tool for \(1979-1935=44\). Notably, even when the generated response includes extra explanation and fails strict exact match, the answer remains semantically correct and is recovered by our LLM-as-a-Judge protocol.

\paragraph{Takeaway.}
Together, these three traces isolate two dominant failure modes for current omni-modal agents: (i) under-calling (insufficient evidence acquisition) and (ii) tool-query drift (evidence acquisition guided by an incorrect prior). The successful trace suggests a practical mitigation: location-first grounding + hypothesis testing + verification before computation, followed by strict answer normalization.

\end{document}